\pdfoutput=1
\documentclass[11pt]{article}

\usepackage[final]{acl}

\usepackage{times}
\usepackage{latexsym}
\usepackage{arydshln} 
\usepackage[T1]{fontenc}
\usepackage[utf8]{inputenc}

\usepackage{microtype}

\usepackage{inconsolata}

\usepackage{graphicx}
\usepackage{booktabs}
\usepackage{makecell}
\usepackage{xspace,mfirstuc,tabulary}
\usepackage{tabularray}
\UseTblrLibrary{booktabs}
\definecolor{c_orange}{HTML}{F89151}

\definecolor{c_purple}{HTML}{BD92BC}
\definecolor{c_blue}{HTML}{6EB4FD}
\definecolor{c_green}{HTML}{99C893}
\newcommand{\datasetname}{\texttt{StorySparkQA}\xspace}
\newcommand{\fairytaleqa}{\textsc{FairytaleQA}\xspace}

\title{\datasetname: Expert-Annotated QA Pairs with Real-World Knowledge for Children's Story-Based Learning}
\author{Jiaju Chen\textsuperscript{*} \\
  East China Normal University \\
  \And
  Yuxuan Lu \\
  Northeastern University \\
  \And
  Shao Zhang\thanks{Work done when visiting Northeastern University.} \\
  Shanghai Jiao Tong University \\
  \AND
  Bingsheng Yao \\
  Northeastern University\\ 
  \And
  Yuanzhe Dong \\
  Stanford University \\ 
  \And
  Ying Xu \\
  Harvard University \\ 
  \AND
  Yunyao Li \\
  Adobe \\ 
  \And
  Qianwen Wang \\
  University of Minnesota\\ 
  Twin Cities \\
  \And
  Dakuo Wang \\
  Northeastern University \\ 
  \AND
  Yuling Sun\thanks{~Corresponding Author: \href{mailto:ylsun@cs.ecnu.edu.cn}{ylsun@cs.ecnu.edu.cn}. } \\
  Lab of Artificial Intelligence for Education\\East China Normal University
  }

\begin{document}
\maketitle
\begin{abstract}
Interactive story reading is common in early childhood education, where teachers expect to teach both language skills and real-world knowledge beyond the story. 
While many story reading systems have been developed for this activity, they often fail to infuse real-world knowledge into the conversation.
This limitation can be attributed to the existing question-answering (QA) datasets used for children's education, upon which the systems are built, failing to capture the nuances of how education experts think when conducting interactive story reading activities.
% This limitation can be attributed to the existing question-answering (QA) datasets used for children's education, upon which the systems are built. These datasets primarily focus on the knowledge within the story content but fail to capture the nuances of how education experts think when conducting interactive story reading activities.
% Parents often struggle with the storytelling activity due to the lack of educational expertise, but it is highly difficult to effectively and exhaustively collect education experts' thinking process and QA pair annotations.
% To bridge this gap, we first designed an annotation framework empowered by real-world knowledge graph to facilitate experts' annotations while collecting their thinking process.
% Further, we leveraged this annotation framework to build \datasetname, a dataset of $5,868$ expert-annotated QA pairs with real-world knowledge beyond story context.
To bridge this gap, we design an annotation framework, empowered by existing knowledge graph to capture experts' annotations and thinking process, and leverage this framework to construct \datasetname dataset, which comprises $5,868$ expert-annotated QA pairs with real-world knowledge.
% that integrate storybook content with an external real-world knowledge graph.
% We designed an annotation workflow supported by a real-world knowledge graph to facilitate education experts creating QA pairs that are appropriate for children's education during storytelling.
% We recruited domain experts in children's education and designed an annotation user interface and a workflow to facilitate them in developing the dataset that has appropriate language and knowledge to teach young children during interactive storytelling.
We conduct automated and human expert evaluations across various QA pair generation settings to demonstrate that our \datasetname can effectively support models in generating QA pairs that target real-world knowledge beyond story content. 
\datasetname\footnote{The \datasetname dataset and annotation framework can also be found in: \url{https://github.com/neuhai/StorySparkQA}} is available at \url{https://huggingface.co/datasets/NEU-HAI/StorySparkQA}.

% A comprehensive benchmark experiment, including automated and human expert evaluation in various QA pair generation (QAG) settings, demonstrates the usability of our \datasetname to support models to generate QA pairs enriched with real-world knowledge beyond story context. 
% Notably, a compact model fine-tuned on \datasetname can reliably outperform robust LLMs, highlighting the complexity of this real-world tasks of interactive story reading. 

% that the T5-Large model fine-tuned on our \datasetname dataset reliably outperforms generic large language models (LLMs) such as GPT-4 and Llama 2 on Rouge-L and Sentence-BERT score.
% Our human evaluation with educational experts further shows that the fine-tuned T5 exhibits superior performance than GPT-4 on the QAG task, only behind human experts, demonstrating the usability and appropriateness of our dataset.
% suggests that human experts still have more nuanced knowledge in the children's education domain than generic LLMs.
\end{abstract}

\section{Introduction}

\begin{figure}[t]
    \centering
    \includegraphics[width=.9\linewidth]{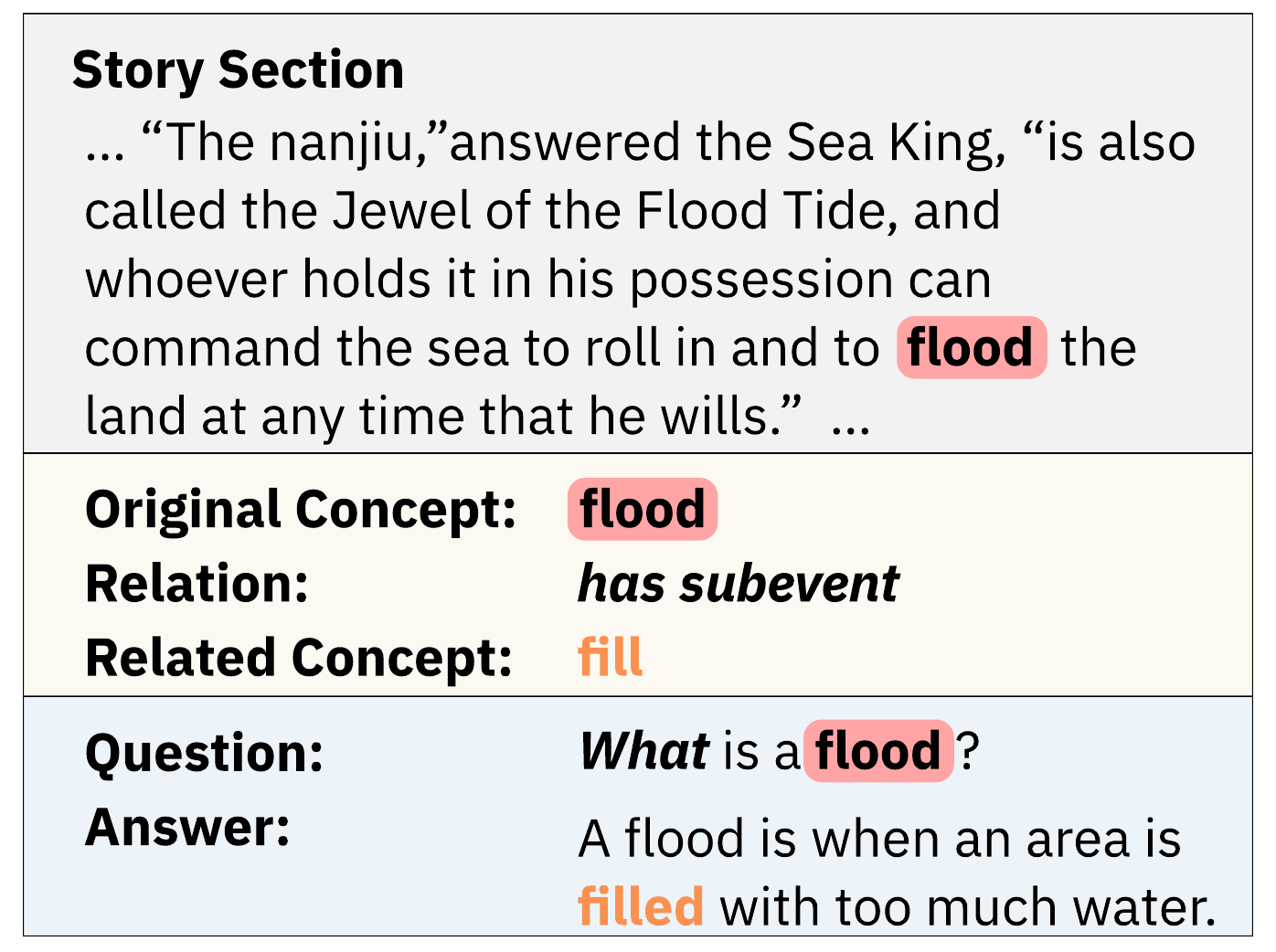}
    \caption{An example of \datasetname dataset.
    In each story section, educational experts select a concept word, link it to a desired external real-world knowledge, and write an appropriate QA pair. Additional data examples of \datasetname are presented in Appendix~\ref{app: sample}.}
    \label{fig:data_eg}
    \vspace{-\baselineskip}
\end{figure}

\textbf{Interactive story reading} is common in early childhood education, where teachers often sit together with preschool children, read storybooks, and proactively engage in question-answering (QA) conversations with them~\cite{wrightStorytellingChildren1995, isbellEffectsStorytellingStory2004}.
Such guided conversations are typically grounded in but beyond the story narratives~\cite{kotamanImpactsDialogicalStorybook2013}, with teachers' expectations of guiding children to learn real-world knowledge and improving their historical, cultural, and emotional awareness~\cite{sunExploringParentNeeds2024c}. 
This immersive story-based interaction has been proven to be effective in better supporting preschooler knowledge learning~\cite{mathemyth}, enhancing their reading comprehension capabilities~\cite{xuSameBenefitsDifferent2021b}, etc. 

Despite the benefits of interactive story reading, teachers often struggle to appropriately conduct such interactive story reading with children because of multi-facet difficulties~\cite{golinkoffLanguageMattersDenying2019a, sunExploringParentNeeds2024c}. Specifically, such interactive story reading needs teachers to identify the knowledge of interest during storytelling, formulate the real-world knowledge piece they want to teach in mind (\textbf{"what to ask"}), then ask an engaging question (\textbf{"how to ask"}) to children at the appropriate time (\textbf{"when to ask"}). 
In home settings, most parents also lack the educational expertise necessary to guide such educational conversations~\cite{golinkoffLanguageMattersDenying2019a, sunExploringParentNeeds2024c}.
Meanwhile, today's parents often hardly maintain constant focus on their children due to the need to deal with other work and family chores at the same time~\cite{zhangStoryBuddyHumanAICollaborative2022a, sunExploringParentNeeds2024c}.

Recently, AI-assisted storytelling systems (e.g. StoryBuddy~\cite{zhangStoryBuddyHumanAICollaborative2022a}, TaleMate~\cite{vargas-diazTaleMateCollaboratingVoice2023}, MatheMyths~\cite{mathemyth}), have demonstrated utility in children's storytelling scenarios~\cite{dietzStoryCoderTeachingComputational2021a}.
These systems primarily utilize the verbal communication interface and advanced language models to support natural conversation with humans~\cite{mahmood2023llm, chan2024human, yang2024talk2care, dietzStoryCoderTeachingComputational2021a}. 
% backed by advanced language models that can drive natural conversation with humans, have demonstrated effectiveness in children's storytelling scenarios~\cite{dietzStoryCoderTeachingComputational2021a}.  
% , with the promises of simulating intelligent interactive storytelling activities with children \cite{dietzStoryCoderTeachingComputational2021a}, supporting their story-based knowledge learning~\cite{mathemyth} and so on. 
% Existing literature has demonstrated the usability and effectiveness of such kind of systems in, for instance, enhancing children's reading comprehension capabilities~\cite{} and knowledge learning~\cite{}, etc.
Nevertheless, existing AI-assisted storytelling systems are not without limitations.
Particularly, building on top of data resources with mostly extractive QA pairs (e.g., FairytaleQA~\cite{xuFantasticQuestionsWhere2022c}) -- where the answers can be found directly in the story narrative -- these systems fall short in helping teaching real-world knowledge beyond the story narrative~\cite{yaoItAITurn2021, zhao2022educational}, which is one of the main expectations of parents and teachers~\cite{sunExploringParentNeeds2024c}. 
% Such limitations significantly impede the practical and educational effectiveness of AI-assisted storytelling systems.
% This challenging task for AI-assisted systems, however, is also very difficult for parents, because of the need for complicated mental procedures and education expertise.

We believe a promising approach to bridge this gap is to effectively and exhaustively collect education experts' knowledge, including their step-by-step thinking process as well as the appropriate QA pairs as final artifacts. 
Nevertheless, to the best of our knowledge, no such data resources exist in the domain of children education. 
Further, the collection of such data resources requires annotators to recall a comprehensive and systematic external knowledge range for a given story text, which is challenging even for education experts~\cite{berry2016pedagogical}. 
As a result, this work aims to facilitate experts' large-coverage knowledge collection and data annotation, and build an expert-labeled, large-scale QA dataset to support story-based educational QA generation with tri-fold contributions: 
\begin{itemize}
    \item We designed an annotation framework empowered by ConceptNet\cite{speerConceptNetOpenMultilingual2017}, a knowledge graph (KG) of structured real-world knowledge, to facilitate education experts creating appropriate story-based educational QA pairs, while collecting experts' mental procedures during data annotation. 
    % Compared to existing annotation frameworks (e.g. \cite{}{}{}) which only support annotators to extract sentences from the text, our framework incorporates external, structured knowledge (i.e. ConceptNet\cite{}) to support experts' large-coverage knowledge collection. We believe our annotation framework possesses the potential to be generalized in analogous domain-specific knowledge collection and annotation tasks requiring structured external knowledge~\cite{vrandecicWikidataFreeCollaborative2014, lehmann_dbpedia_2015}, such as clinicians using structured rules and knowledge when diagnosing patients~\cite{elsayed_2_2023, american_diabetes_association_diagnosis_2011}.  
    \item Based on the proposed annotation framework, we build \datasetname, an expert-labeled QA dataset consisting of $5,868$ story-based QA pairs infused with real-world knowledge. 
    %Compared to existing datasets in children education domain (e.g., StoryQA~\cite{Zhao2023}, \textsc{FairytaleQA}~\cite{xuFantasticQuestionsWhere2022c}, and EduQG~\cite{hadifarEduQGMultiFormatMultipleChoice2023a}), which mostly comprise QA-pairs grounded in the story, \datasetname associates with real-world knowledge beyond the story narratives. \datasetname can benefit different research aspects in the children's education domain, particularly in better understanding domain experts' thinking process, training models to generate story-based QA pairs infused with real-world knowledge, with an ultimate goal of broadening children's knowledge scope beyond story narratives that parents expect.    
    \item We demonstrate the utility of our \datasetname on the QA pair generation (QAG) task, benchmarked with a set of popular language models (fine-tuned T5-Large~\cite{T52020}, zero-shot, few-shot, and Chain-of-Thought with GPT-4~\cite{openaiGPT4TechnicalReport2023a}, Llama 2~\cite{touvron_llama_2023}, etc.\footnote{We also experiment with GPT-3.5, Flan-T5-XXL~\cite{chungScalingInstructionFinetunedLanguage2022}, Alpaca-7B~\cite{alpaca} and Mistral-7B~\cite{jiang_mistral_2023}. We report the results in Appendix~\ref{app:FullResult}.}) through automated evaluation and human expert evaluations. 
\end{itemize} 
\datasetname can benefit different research directions in the children education, particularly in better understanding domain experts' thinking process, and training models to generate story-based QA pairs infused with real-world knowledge, with the ultimate goal of broadening children's knowledge scope beyond story narratives that parents and teachers expect.
In addition, we believe our annotation framework has the potential to be generalized to domain-specific tasks analogous to the real world that require structured external knowledge~\cite{vrandecicWikidataFreeCollaborative2014, lehmann_dbpedia_2015}, such that clinicians use structured guidelines and knowledge for diagnosing~\cite{elsayed_2_2023, american_diabetes_association_diagnosis_2011}.

\section{Related Work}

\subsection{Children Education and Real-World Knowledge Resources}

Existing datasets in the education domain (e.g., StoryQA~\cite{Zhao2023}, \textsc{FairytaleQA}~\cite{xuFantasticQuestionsWhere2022c}, and EduQG~\cite{hadifarEduQGMultiFormatMultipleChoice2023a}) mostly comprise QA-pairs grounded in the story, lacking real-world knowledge beyond the story.
We present key properties of related children education datasets in Table~\ref{tab:related_datasets}.
% in Appendix~\ref{EduQADataset}
On the other hand, general-purpose datasets like CommonsenseQA~\cite{talmor2018commonsenseqa} and SciQA~\cite{SciQA2023} integrate crowd-sourced commonsense with narratives, but lack educational appropriateness aligned with children's knowledge level.

% Structured knowledge representation is widely adopted to store real-world knowledge due to its simplicity for use and maintenance.
% However, m
Many popular real-world knowledge resources, such as ATOMIC~\cite{sapATOMICAtlasMachine2019} and Wikidata~\cite{vrandecicWikidataFreeCollaborative2014}, are too complicated for children's knowledge level. 
A more appropriate option is ConceptNet~\cite{speerConceptNetOpenMultilingual2017}, a very large-scale knowledge graph for real-world concepts and relations stored in triples: ($concept_1$, $relation$, $concept_2$).
The simplicity of triple representations makes ConceptNet suitable for children education, as demonstrated in prior literature~\cite{xuFusingContextKnowledge2020}, thus, our work also leverages ConceptNet to support experts' annotation process.

\subsection{QA Pair Annotation Frameworks}
Some existing annotation frameworks, such as Potato~\cite{pei_potato_2022} and Piaf~\cite{keraron_project_2020}, mostly focus on facilitating extractive QA pairs grounded in the text, that is, providing source texts and allowing annotators to highlight a span of text as an answer to a question. Some others, like the annotation toolkit for StoryQA~\cite{Zhao2023}, support free-form input, allowing annotators to type in answers in their own words through the data collection user interface. In either type, existing annotation frameworks can't support story-based external knowledge collection and story data annotation effectively, in which annotators are required to recall comprehensive and systematical real-world knowledge for a given story text. 
Our study bridges this gap by proposing an external knowledge-empowered annotation framework.

% \yuling{
%  Leveraging external knowledge (e.g. ATOMIC~\cite{sapATOMICAtlasMachine2019} and Wikidata~\cite{vrandecicWikidataFreeCollaborative2014}) to construct external knowledge-infused datasets has been widely adopted~\cite{talmor2018commonsenseqa, SciQA2023}. Considering children's knowledge and cognitive level, we leverage ConceptNet~\cite{speerConceptNetOpenMultilingual2017}, a large-scale knowledge graph with real-world concepts and relations stored in simple triples: ($concept_1$, $relation$, $concept_2$), to support annotation process.
% }

% Existing annotation frameworks either focus on extractive QA pairs (e.g., Potato~\cite{pei_potato_2022} and Piaf~\cite{keraron_project_2020}), which only allows annotators to highlight text spans, or supports free-form input but provide no domain-specific knowledge assistance~\cite{Zhao2023}. 
% These frameworks
% Thus, it is essential to facilitate experts' annotation with an annotation workflow that assists experts in identifying key elements within narratives, presents them with coherent and structured external knowledge, and supports them in writing QA pairs based on the retrieved knowledge.

\subsection{QA Pair Generation (QAG)}
Fine-tuning traditional pre-trained language models like BERT~\cite{devlinBERTPretrainingDeep2019} on QAG datasets for end-to-end generation was a prevalent approach, but such methodology heavily depends on the training data quality and lacks control of generated content, which is inappropriate for the children education domain.
Existing works also attempted to design multi-step generation pipelines, which offer better control of the generated content~\cite{yaoItAITurn2021, wan2024sciqag}.

Recent advancement in large language models (LLMs), such as GPT-3.5, GPT-4~\cite{openaiGPT4TechnicalReport2023a}, and Llama 3~\cite{llama3}, supports free-form natural language input and output without the need for tuning model parameters. 
Many prompting strategies were developed to further enhance models' task-solving and domain-adaptation capabilities, including few-shot in-context learning (i.e., add a few examples in input)~\cite{brownLanguageModelsAre2020, yao2024more}, Chain-of-Thought (i.e., ask models to think ``step-by-step'')~\cite{weiChainThoughtPrompting2022}, etc.
The performance of these prompting methods on the general QAG task has also been evaluated \cite{lingAutomaticQuestionanswerPairs2024, luHumanStillWins2023a}.
However, to what extent these prompting and modeling strategies are effective in \textbf{the QAG task for knowledge beyond the story content}, and whether a compact language model fine-tuned on domain-specific datasets performs better or worse than generic LLMs in the context of children education remains underexplored.
For example, recent work demonstrates the unreliability of generic LLMs for the classification of mental health issues and the superiority of fine-tuning domain-specific language models with high-quality datasets~\cite{xu2024mental}. 
Our work attempts to step forward through comprehensive evaluation in Section~\ref{sec:eval}.

% With the rapid development of pre-trained language models~\cite{devlinBERTPretrainingDeep2019, liuRoBERTaRobustlyOptimized2019a}, neural network-based methods~\cite{zhouNeuralQuestionGeneration2018a, zhaoEducationalQuestionGeneration2022b} are more prevalent recently. 
% Yet, the generation qualities of neural network-based approaches highly depend on the training datasets.
% There remains a lack of datasets containing external domain knowledge to evaluate models' QAG performance for domain-specific tasks, such as interactive storytelling in children's education.

% Recent advances in LLMs~\cite{chungScalingInstructionFinetunedLanguage2022, openaiGPT4TechnicalReport2023a} show exceptional natural language generation (NLG) capabilities. 
% While conversational LLMs like GPT-4, and FLAN-T5 demonstrate superior zero-shot and few-shot in-context learning (ICL)~\cite{ICL2020}, Chain-of-Thought (CoT)~\cite{CoT2024} performance, their adaptability and performance in specialized domains, such as children's education, remain underexplored.
% We experiment with a series of QAG pipelines using SoTA LLMs to assess their performance thoroughly.

\section{Expert Annotation Framework}
\label{sec:dataset_task}
\begin{figure}[!tp]
    \centering
    \includegraphics[width=.99\linewidth]{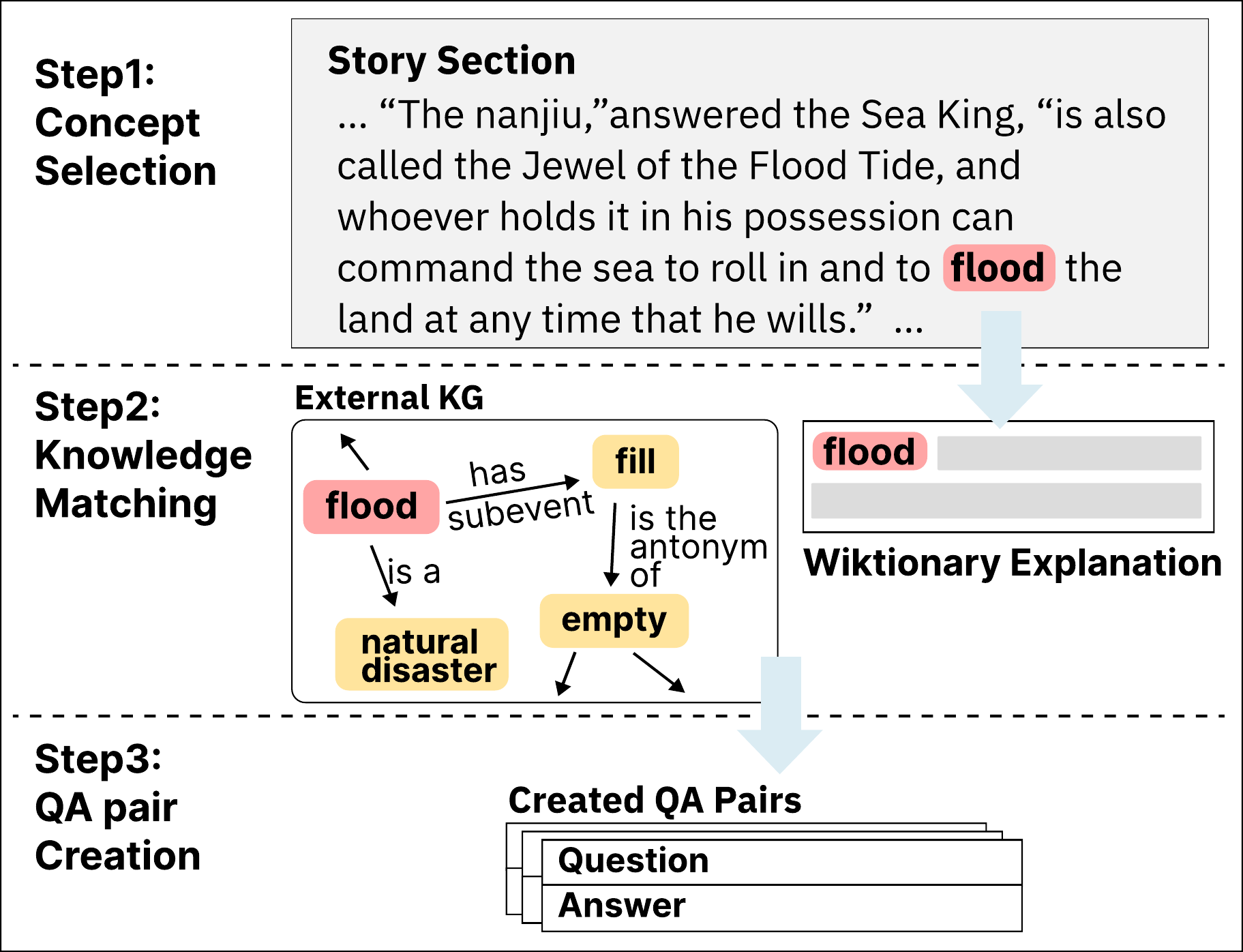}
    \caption{Workflow of the experts’ annotation process.
    Experts need to select a concept first, then match it with the most suitable knowledge, and finally create a QA pair based on the selected knowledge.}
    \label{fig:workflow}
    \vspace{-\baselineskip}
\end{figure}

To better understand experts' mental procedures during annotating QA pairs enriched with external real-world knowledge, we proposed a three-step QA pair annotation framework with interactive user interfaces (UI). Particularly, considering the challenges facing annotators in recalling the comprehensive and systematical external knowledge for a given story text~\cite{berry2016pedagogical}, our framework incorporates ConceptNet, a large-scale real-world Knowledge Graph, to support experts' large-coverage knowledge collection. % We present 
The workflow of our annotation framework is shown in Figure~\ref{fig:workflow}. 

% \begin{enumerate}
%     \item \textbf{Concept Selection}: Experts identify an educational-appropriate concept from the story for children.
%     \item \textbf{Knowledge Matching}: Experts select a real-world knowledge triple, retrieved and recommended from ConceptNet, based on the identified concept.
%     This step connects story content with external real-world knowledge.
%     \item \textbf{QA pair Creation}: Experts write an educational-appropriate QA pair based on the selected triple.
% \end{enumerate}

% To facilitate the annotation process by providing recommendations for external real-world knowledge, we design our annotation framework by retrieving and recommending real-world knowledge triples from ConceptNet~\cite{speerConceptNetOpenMultilingual2017}, a publicly available, large-scale real-world Knowledge Graph.

% \vspace{-5pt}
\begin{figure*}[ht]
    \centering
    \includegraphics[width=.88\linewidth]{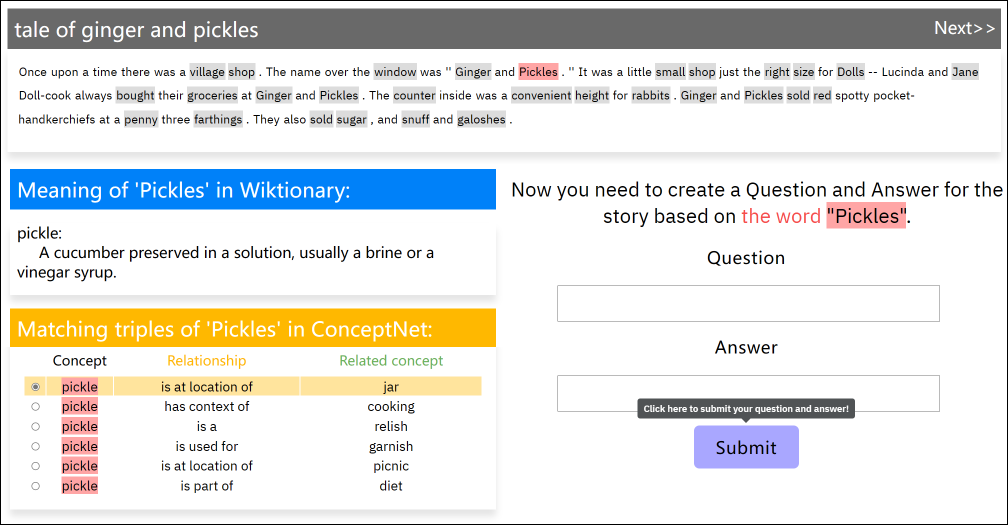}
    \caption{The user interface to facilitate our annotation task. The words highlighted in grey are candidate concepts. The blue block shows the Wiktionary explanation, and the yellow block lists our recommended triples.}
    \label{fig:ui_QA_creation}
    \vspace{-\baselineskip}
\end{figure*}

\paragraph{Step 1. Concept Selection}
\label{sec:dataset_framework_preprocessing}
In this step (UI shown in Figure~\ref{fig:section}), experts identify an educationally suitable concept from the story text. 
We develop a collection of heuristics to filter candidate concepts that are tier 1 or tier 2~\footnote{Tier 1 words are common and basic words. Tier 2 contains high-frequency words of various domains~\cite{beck2013bringing}.} vocabulary and a concrete noun, verb, or adjective.
First, we leverage the spaCy~\cite{spacy2} English model to filter auxiliary words and punctuation~\footnote{tagged by `auxiliary', `adposition', `determiner', `particle', `punctuation', `symbol', and `other'} from the original story text.  
Then, we use AllenNLP's~\cite{Gardner2017AllenNLP} semantic role labeling tool to tag the latent structure of each sentence in the story context.
This process identifies and retains key elements represented by semantic roles, including agents, goals, and results, which are subsequently treated as potential candidate concepts.
% We design the UI for step 1 to display one story section and allow experts to select highlighted candidate concepts in grey, as shown in Figure~\ref{fig:section}.
% and candidate concepts are highlighted in grey to select. 

\paragraph{Step 2. Knowledge Matching}
\label{sec:dataset_framework_retrieval}
This step (UI shown in Figure~\ref{fig:triples}) allows experts to select real-world knowledge based on the concept selected previously.
Inspired by ~\citet{xuFusingContextKnowledge2020}'s work of combining and filtering knowledge from Wiktionary~\footnote{\url{https://www.wiktionary.org/}} and ConceptNet~\cite{speerConceptNetOpenMultilingual2017} for commonsense question answering, we implement a knowledge matching module that can retrieve and rank external knowledge associated with each concept selected by the experts.

Specifically, once experts select a candidate concept, our knowledge matching module 1) retrieves a list of real-world knowledge triples, with the format of \textit{(source concept, relation, target concept)} from ConceptNet; 2) filters out weak relations in ConceptNet
% , leaving main relation types for annotation 
(complete relation list in Appendix~\ref{app:filter_relations}), and 3) rank knowledge triples by concatenating concepts and relationships, and calculating the average similarity between every other triple with the Term Frequency-Inverse Document Frequency (TF-IDF)~\cite{ramos2003using}.

% The second step in our knowledge matching module is to rank and select diverse and representative triples from all retrieved real-world knowledge triples associated with the selected concept.
% We use the concatenation of the relation and related concept in each triple to calculate the average similarity between every other retrieved triple using the Term Frequency-Inverse Document Frequency (TF-IDF). 

We rank all retrieved triples with $1 - \overline{s} + w$, where $\overline{s}$ denotes the similarity score and $w$ denotes the weight of a triple provided by ConceptNet, reflecting the combined influence and credibility of the triple by summing up the weights coming from all the sources that support it. 
The top six ranked triples are shown to annotators to balance between providing a sufficient selection and avoiding excessive distractions during annotation.
We also retrieve the explanation for concepts from Wiktionary to better facilitate experts' annotations.
% The UI for this step is designed to display a list of recommended real-world knowledge triples along with the Wiktionary explanation to facilitate experts' selection (Figure~\ref{fig:triples}).
% Leveraging our knowledge matching module, we develop the second interface  (Figure~\ref{fig:triples}) by retrieving and matching associated structured knowledge (i.e., real-world knowledge triples) with the selected concept.

\paragraph{Step 3. QA pair Annotation}
\label{sec:dataset_framework_annotation}
This step enables annotators to create a QA pair based on the real-world knowledge triple they selected in step 2, and the corresponding UI is shown in Figure~\ref{fig:ui_QA_creation}.
In this step, experts are instructed to incorporate one concept in the question or answer and include the relation from the triple in the resulting QA pair.

\section{\datasetname}

\begin{table*}[!htb]
\centering
\resizebox{\linewidth}{!}{
\begin{booktabs}{
    colspec={l|cccc|cccc|cccc},
    cells={valign=m},
    width=\linewidth,
    row{1}={font=\bfseries},
    row{3}={font=\bfseries\small},
    cell{1}{1}={r=3}{c,m},
    cell{1}{2,6,10}={c=4}{c},
    cell{2}{2,6,10}={c=4}{c},
    cell{4-10}{2-13}={font=\small},
}
    \toprule
    \datasetname & Train &&&& Validation &&&& Test \\
     & $232$ books with $4, 300$ QA pairs &&&& $23$ books with $769$ QA pairs &&&& $23$ books with $799$ QA pairs \\ 
     % \cmidrule[lr]{2-5} \cmidrule[lr]{6-9} \cmidrule[lr]{10-13}
     & Mean & SD & Min & Max & Mean & SD & Min & Max & Mean & SD & Min & Max \\
    \midrule
    \midrule
     {\# sections / story} & $14.4$ & $8.8$ & $2$ & $60$ & $16.5$ & $10.0$ & $4$ & $43$ & $15.8$ & $10.8$ & $2$ & $55$ \\
     {\# tokens per story} & $2160.9$ & $1375.9$ & $228$ & $7577$ & $2441.8$ & $1696.9$ & $425$ & $5865$ & $2313.4$ & $1369.6$ & $332$ & $6330$ \\
     {\# tokens / section} & $149.6$ & $64.8$ & $12$ & $447$ & $147.8$ & $56.7$ & $33$ & $298$ & $145.8$ & $58.6$ & $24$ & $290$ \\
     {\# questions / story} & $18.5$ & $14.5$ & $2$ & $126$ & $33.4$ & $22.1$ & $4$ & $115$ & $34.7$ & $21.1$ & $8$ & $90$ \\
     {\# questions / section}& $1.3$ & $0.6$ & $1$ & $9$ & $2.1$ & $0.3$ & $2$ & $3$ & $2.1$ & $0.3$ & $2$ & $3$ \\
     {\# tokens / question}& $5.2$ & $2.0$ & $3$ & $19$ & $5.9$ & $1.6$ & $3$ & $13$ & $6.0$ & $1.7$ & $3$ & $13$\\
     {\# tokens / answer}& $5.4$ & $3.7$ & $1$ & $20$ & $3.8$ & $2.3$ & $1$ & $12$ & $3.8$ & $2.3$ & $1$ & $12$ \\
    \bottomrule
\end{booktabs}
}
\caption{
 Core statistics of our \datasetname dataset, which has $278$ books and $5, 868$ QA pairs.}
\label{tab:dataset_statistics}
\vspace{-\baselineskip}
\end{table*}

\datasetname aims to facilitate teachers' interactive story reading with appropriate real-world knowledge: \textbf{practical, factual, everyday information that helps preschoolers understand the world around them.} 
Our dataset consists of $5,868$ QA pairs annotated by children education experts leveraging our designed annotation framework.
We present the core statistics of \datasetname in Table~\ref{tab:dataset_statistics} and show one example in Figure~\ref{fig:data_eg}.

\subsection{Source Narrative}
Among the existing story-based datasets for children education, \fairytaleqa~\cite{xuFantasticQuestionsWhere2022c} comprises $278$ classic fairytale stories of various origins, and all the stories have been evaluated as suitable for 8\textsuperscript{th}-grade children and younger.
The original stories were parsed by education experts into shorter sections of around $150$ words, which leads the \fairytaleqa dataset to a unique and high-quality text corpus for children's reading comprehension.
As a result, we take the story sections from \fairytaleqa as the source text for our \datasetname dataset.

\subsection{Annotation Process}

Following our annotation framework, we recruit 11 children education experts for the annotation task. 
The experts all have a minimum of 3 years of practical experience (e.g., kindergarten teachers) in learning science and possess relevant educational backgrounds.
% To ensure the created QA pairs and retrieved knowledge suit parents' real-world needs, experts are asked to mimic parents' habits during storytelling. 
For each story section, experts are asked to first identify a concept from the story by selecting concepts that are most beneficial for children's education from story text.
The experts then proceed to select a real-world knowledge triple associated with the selected concept and create a QA pair based on the selected triple.
In this process, experts are asked to consider children's cognitive and knowledge levels and write QA pairs that are most appropriate for 3- to 6-year-olds.
We collect experts' mental procedures by recording their selected concepts, real-world knowledge triples, and created QA pairs during the annotation process.
% Aligned with the user interface design demonstrated in Figure~\ref{fig:ui_QA_creation}, we present the 3-step workflow of QA pair annotation below, which follows Figure~\ref{fig:RAG}.

\subsubsection{Cross-Validation}
\label{sec:dataset_validation}

To ensure the quality and consistency of annotated QA pairs among annotators, as well as to evaluate agreement in selecting triples and creating QA pairs between annotators, we designed additional cross-validation procedures with corresponding UIs.
We randomly selected $50$ QA pairs in both the test and validation split ($100$ QA pairs in total) and two annotators were asked to cross-validate each other's annotation (denoted by $annotator_A$ and $annotator_B$, accordingly):
\begin{enumerate}
\item Shown in Figure~\ref{fig:val-section}, $annotator_A$ is provided with the story section and the concept selected by $annotator_B$.
For each selected concept, $annotator_A$ is asked to rank the top 3 triples from the same recommended triple list given to $annotator_B$, verifying the triple selection agreement between annotators (Figure~\ref{fig:val-rank}).
\item In the next step, $annotator_A$ is asked to create an QA pair based on the word and triple selected by $annotator_B$, evaluating the similarity of QA pairs between annotators given the identical triple (Figure~\ref{fig:val-QA}).
\item After submitting the QA pair in Step $2$, $annotator_A$ is provided with the question created by $annotator_B$ based on the same triple, and $annotator_A$ is asked to write an answer to the question to cross-validate the question-answering agreement (Figure~\ref{fig:val-answer}).
\end{enumerate}

% \subsubsection{Cross-Validation Results}
% \label{sec:dataset_validation_result}

Of the $100$ randomly selected sections in the validation and test splits, $86\%$ of the triples that appear in the top-3 list are selected by both annotators, and $56\%$ of the triples are ranked top by the validator, indicating a very high consistency between experts for triple selection.
In addition, we evaluate the similarity of the concatenated QA pairs created by each of the annotators based on the same triple with Rouge-L~\cite{linROUGEPackageAutomatic2004a} and SBERT~\cite{reimersSentenceBERTSentenceEmbeddings2019b} scores.
The Rouge-L F1 score of QA pair creation between annotators is 0.53, and the SBERT score is 0.80, showing a shared tendency among experts when selecting real-world knowledge and creating a QA pair that is both beneficial and suitable for children education.
% This observation reinforces the necessity of experts' annotation in constructing a high-quality QA dataset for children's education.

\begin{figure}[!tp]
    \centering
    \includegraphics[width=.9\columnwidth]{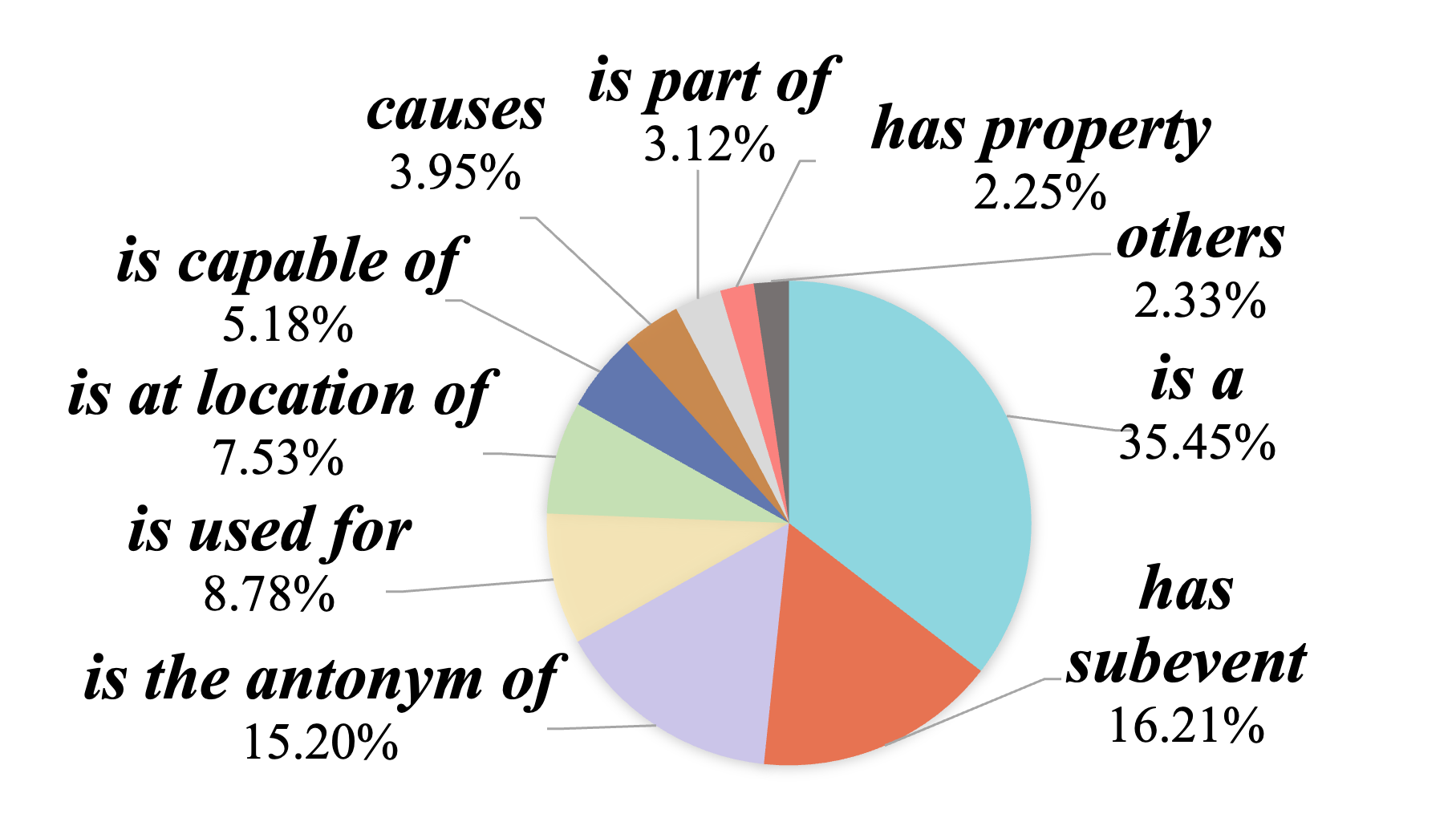}
    \caption{Distribution of real-world knowledge relations annotated by experts in the \datasetname dataset}
    \label{fig:rel_distribution}
    \vspace{-\baselineskip}
\end{figure}

\subsection{Statistics and Analysis of \datasetname}
\label{sec:dataset_stats}

\paragraph{Statistics of \datasetname}
Figure~\ref{fig:rel_distribution} demonstrates the distribution of real-world knowledge relations in \datasetname, and Table~\ref{tab:dataset_statistics} illustrates detailed statistics of the dataset.
On average, each section is annotated with approximately 1.4 QA pairs. In \datasetname, the top 3 real-world knowledge relations selected by experts are \textit{``is a''}, \textit{``has subevent''} and \textit{``is the antonym of''}, respectively constituting $35.5$\%, $16.2$\% and $15.2$\% of all real-world knowledge relations. 
%\textit{is used for}, \textit{is at location of} and \textit{is capable of} each constitute $8.8$\%, $7.5$\%, and $5.2$\% of all commonsense relations. The proportion of other relations is less than $5$\%. 
The distribution of question types in \datasetname is shown in Table~\ref{tab:distribution} in Appendix~\ref{app:distribution}. 
In \datasetname, questions starting with ``what'' are the most common type of question, constituting $86.0$\%. % of all questions. 
Questions starting with ``why'' and ``how'' constitute about $7.2$\% and $2.4$\%, respectively. 

\paragraph{Analysis of \datasetname's Alignment with Real-World Needs}
According to experts' annotations, real-world knowledge relation \textit{``is a''} and questions start with ``what'' have a much higher proportion than the others. 
Moreover, of all experts' annotations, the selected concept words are mostly nouns (65.06\%) and adjectives (22.26\%), which are easy for children to comprehend.
Considering the cognitive development of children, especially those aged 3 to 6, they are typically in the stage of language development and exploration, full of curiosity about the world~\cite{chouinard_childrens_2007, jirout_childrens_2012}. It is therefore natural for them to ask questions as a way to satisfy this curiosity.
Consequently, teachers are more inclined to use ``what'' questions with simple vocabulary to inspire children's thinking and encourage them to actively acquire knowledge~\cite{taylor1994children, yu_pedagogical_2019}. 
Consistent with the actual habits of teachers, experts' annotated questions have a high consensus that ``what'' questions are more aligned with children's cognitive characteristics.

% comparison with existing datasets
\paragraph{Comparison with Existing Datasets}
Compared with existing QA datasets for children education~\cite{xuFantasticQuestionsWhere2022c, Zhao2023}, \datasetname is unique mainly in the annotation process and data composition.
Going beyond direct QA pair annotation utilized by most QA datasets (e.g., \fairytaleqa and StoryQA), \datasetname's annotation process provides step-by-step support with structured real-world knowledge to ease the effort that the experts need to craft children-appropriate and knowledgeable QA pairs.
In addition, \datasetname includes not only QA pairs, but also corresponding story texts and expert-selected real-world knowledge triples, which reflects experts' mental procedures when creating these children-appropriated, knowledgeable QA pairs and leads to a more comprehensive dataset for children's knowledge expansion. Catering to teachers' practical needs for knowledge expansion in interactive story reading, \datasetname complements existing QA datasets, which often focus on narrative comprehension and commonsense question-answering.
The key properties of \datasetname and related children education datasets are illustrated in Table~\ref{tab:related_datasets} in Appendix~\ref{EduQADataset}.
% Complementing existing datasets for other aspects of children's education (e.g., narrative comprehension)

\section{Benchmark Experiment}
\label{sec:eval}

We benchmark the quality and usability of our \datasetname on the QAG task, which is required to meet the needs of teachers to guide children to learn some real-work knowledge during practical interactive story reading, as well as existing work of developing AI-assisted storytelling and reading systems~\cite{yaoItAITurn2021, dietzStoryCoderTeachingComputational2021a, zhangStoryBuddyHumanAICollaborative2022a}. 
We conduct an \textbf{automated evaluation}, reported in Section~\ref{sec:auto_eval} to measure the semantic similarity of generated QA pairs with experts-annotated QA pairs, benchmarked with a T5-Large model fine-tuned on \datasetname and a set of robust LLMs. 
Considering the limitation of automated evaluation in evaluating the educational appropriateness of generated QA pairs, we further conduct a \textbf{human evaluation}, reported in Section~\ref{sec:human_eval}, with children education experts. 

% Catering to parents' practical needs to ask questions with external real-world knowledge during the interactive storytelling activity, it is a natural and critical task for existing interactive storytelling models~\cite{yaoItAITurn2021} to ask questions and provide answers simultaneously.
% Therefore, we assess the utility of our \datasetname on the question-answer generation (QAG) task and present the benchmark experiments in this section.

% We evaluate the \datasetname's effectiveness in enabling models to generate QA pairs suitable for children's education by 
% fine-tuning a T5-Large model with \datasetname dataset and evaluating its QA pair generation (QAG) performance compared with State-of-The-Art (SoTA) LLMs. 

% Apart from automatic evaluation, which we believe can not faithfully represent the domain specialty with a generic evaluation model, we conduct a \textbf{human evaluation} to further assess the quality of LLM-generated QA pairs and expert annotations from an educational perspective.

\subsection{Automated Evaluation}
\label{sec:auto_eval}
We now elaborate on the settings and results of our QAG experiments with various language models, through which to demonstrate the usability of \datasetname.
% The experiment settings and results are presented in the following subsections.

\subsubsection{Experiment Settings}

The QAG task involves taking a story section as input and generating the QA pairs.
To exploit LLMs' comprehensive generation ability, we design two variations to simulate experts' annotation process: 
\begin{enumerate}
    \item \textit{QA pair generation}: Generate the QA pair.
    \item \textit{QA pair and triple generation}: Generate the associated real-world knowledge triple alongside the QA pair.
    % to mimic experts' annotation process.
\end{enumerate}

The automatic evaluation comprises six popular LLMs: GPT-3.5, GPT-4~\cite{openaiGPT4TechnicalReport2023a}, FLAN-T5-XXL~\cite{chungScalingInstructionFinetunedLanguage2022}, Alpaca-7B~\cite{alpaca}, Mistral-7B~\cite{jiang_mistral_2023} and Llama 2(7B)~\cite{touvron_llama_2023}.
We carefully design the prompt inputs (Appendix~\ref{app: gpt_prompts}) with clear and informative instructions, including 13 relation types (Appendix~\ref{app:filter_relations}) in ConceptNet.
The goal is to leverage LLMs to generate diverse triples similar to those created by human education experts.

For each LLM involved in this experiment (GPT-3.5, GPT-4, FLAN-T5-XXL, Alpaca, Mistral, and Llama 2), we employ \textbf{zero-shot}, \textbf{few-shot in-context learning (ICL)}~\cite{brownLanguageModelsAre2020} approaches to thoroughly examine the QAG performance of these models with different prompting strategies. 
For GPT-3.5 and GPT-4, we also use \textbf{Chain-of-Thought (CoT)}~\cite{weiChainThoughtPrompting2022} to further explore their QAG capabilities.
Randomly sampled examples from the validation split are used as demonstrations for the few-shot ICL approaches.
We also fine-tune a T5-Large model to examine how a much smaller domain-specific model, supported by expert-annotated triples as additional input, performs compared to generic LLMs. 
The experiment settings and hyper-parameters are reported in Appendix \ref{app:hyper}.

We utilize \textbf{Rouge-L}~\cite{linROUGEPackageAutomatic2004a} to evaluate the quality of concatenated QA pairs between the generated ones and two expert-annotated ground truths of each data, and report the averaged score across all data in the test split.
For the setting that generates triples along with QA pairs, we evaluate the generated triples and QA pairs separately.
The Rouge-L F1 score is chosen because it captures the sequential matching of texts, reflecting whether the generated QA pairs are textually similar to experts’ annotation. 
However, we are aware that Rouge-L is not without limitations, such as a lack of semantic understanding. To address this, we incorporate~\textbf{SBERT} using Sentence Transformer~\cite{reimersSentenceBERTSentenceEmbeddings2019b} for automated evaluation and present the full results in Appendix~\ref{app:FullResult}. We also conduct a human evaluation (see in Section~\ref{sec:human_eval}) to further assess the quality of generated QA pairs.
We perform experiments with GPT-3.5 and GPT-4 three times for each setting to calculate a robust and reliable average score.

\subsubsection{Results and Analysis}

\begin{table}[t!]
    \centering
    \resizebox{0.99\linewidth}{!}{
    \begin{booktabs}{
    colspec={cl|cc},
    row{1}={font=\bfseries\small},
    width=\linewidth,
    cells={m},
}
\toprule[1pt]
Model & {Prompting\\Strategy} & {QAG\\w/o Triples} & {QAG\\w/ Triples} \\
\midrule
\midrule%[1pt]
{T5-Large\\Fine-Tuned\\(0.77B)} & - & \textbf{0.332} & \textbf{0.279} \\
\cmidrule[lr]{1-4}
\SetCell[r=2]{c}{Alpaca\\(7B)} & zero-shot & 0.124 & 0.266 \\
& few-shot & 0.251  & 0.239  \\
\cmidrule[lr]{1-6}
\SetCell[r=2]{c}{Mistral\\(7B)} & zero-shot & 0.229  & 0.209 \\
& few-shot & 0.267 & 0.257 \\
\cmidrule[lr]{1-6}
\SetCell[r=3]{c}{Llama 2\\(7B)} & zero-shot & 0.213 & 0.177 \\
& 1-shot & 0.192 & 0.206 \\
& 5-shot & 0.241 & 0.269 \\
\cmidrule[lr]{1-4}
\SetCell[r=4]{c}{GPT-3.5} & zero-shot & 0.194 & 0.220 \\
& 1-shot & 0.239 & 0.252 \\
& 5-shot & 0.262 & 0.264 \\
& CoT & - & 0.259 \\
\cmidrule[lr]{1-4}
\SetCell[r=4]{c}{GPT-4} & zero-shot & 0.277 & 0.243 \\
& 1-shot & 0.272 & 0.251 \\
& 5-shot & 0.287 & 0.248 \\
& CoT & - & 0.262 \\
\bottomrule[1pt]
\end{booktabs}}
\caption{
 QAG performance of LLMs with different prompting strategies and the fine-tuned T5-Large model. \textbf{Bolded numbers} are the best scores within each setting.
 }
\label{tab:exp1_result}
\vspace{-\baselineskip}
\end{table}

In table~\ref{tab:exp1_result}, we show the zero-shot, few-shot ICL, and CoT performances on all models in both settings of the QAG task.

Generally, zero-shot QAG performance on these models falls short of the few-shot ICL QAG performance.
Particularly, Alpaca's few-shot performance when generating QA pairs without triples falls behind the zero-shot setting.
This is due to the repetitive and less diverse outputs generated by the model in the zero-shot setting (e.g., \textit{`What is the relation between A and B? A is the antonym of B.'} ).
Such QA pairs are repetitive across the results, leading to a higher Rouge-L score as the words in it 
have a higher chance to match with the experts’ annotations. Particularly, the relation \textit{`is the antonym of'} occupies a significant proportion in the concatenation of QA pairs.

Remarkably, models using 5-shot demonstrations outperform those using 1-shot demonstrations.
It is notable that GPT-3.5 achieves better performance in the few-shot setting compared to GPT-4.
We believe this is caused by GPT-4's exceptional Natural Language Generation versatility, potentially resulting in longer and more grammatically complicated QA pairs that are unsuitable for 3- to 6-year-olds due to advanced vocabulary and knowledge comprehension challenges~\cite{ouellette2006s, perfetti2014word}.
Meanwhile, models employing the Chain-of-Thought prompting method do not imply an obvious improvement compared to the few-shot ICL QAG performance.

For the setting of generating triples along with QA pairs (\textit{w/ triples}), the results do not indicate an improvement in QAG through the step of generating real-world knowledge triples.
We attribute this to the potential complexity of the task that asks LLMs to generate real-world knowledge triples and corresponding QA pairs simultaneously.

It is worth noting that T5-Large fine-tuned on our \datasetname has a relatively better performance in generating QA pairs enriched with real-world knowledge for children than conversational LLMs like GPT-3.5 and GPT-4 by Rouge-L.
Additional analysis of the generated external knowledge types among experts' annotations, fine-tuned T5-Large, and GPT-4 is demonstrated in Appendix~\ref{app: distribution_analysis}.
% Overall, the QAG pipeline has a relatively limited upper bound in both Rouge-L score and SBERT.

\subsection{Human Evaluation}
\label{sec:human_eval}
\begin{table*}[t!]
    \centering
    % \small
    % \vspace{2\baselineskip}
    \begin{booktabs}{
        colspec={llccccc},
        row{1}={font=\bfseries},
        width=\linewidth,
        % cell{2,7}{1}={r=5}{},
        % row{2-11}={ht=2\baselineskip},
    }
    \toprule
        Dimension & Model & Mean & SD & t & df & p-value\\
        \midrule
        \midrule
        {\SetCell[r=3]{l}{\textbf{Grammar Correctness}}}
        & Human & 4.893 & 0.560 & & & &\\
        & T5-Large Fine-Tuned & 4.842 & 0.585 & 1.259 & 279 & 0.209\\
        & GPT-4 & \textbf{4.871} & 0.514 & 0.646 & 279 & 0.519 \\
        \cmidrule[lr]{1-7}
        {\SetCell[r=3]{l}{\textbf{Answer Relevancy**}}}
        & Human & 4.696 & 0.683 & & & &\\
        & T5-Large Fine-Tuned & 4.329 & 1.111 & 5.487 & 279 & <0.01 \\
        & GPT-4 & \textbf{4.379} & 0.869 & 5.123 & 279 & <0.01 \\
        \cmidrule[lr]{1-7}
        {\SetCell[r=3]{l}{\textbf{Contextual Consistency*}}}
        & Human & 4.657 & 0.882 & & & &\\
        & T5-Large Fine-Tuned & \textbf{4.639} & 0.972 & 5.487 & 279 & 0.729 \\
        & GPT-4 & 4.529 & 0.974 & 2.240 & 279 & 0.026 \\
        \cmidrule[lr]{1-7}
        {\SetCell[r=5]{l}{\textbf{Educational}}}
        & Human & 4.493 & 0.892 & & & & \\
        {\textbf{Appropriateness**}}
        & T5-Large Fine-Tuned & \textbf{4.325} & 0.972 & 2.937 & 279 & <0.01 \\
        & GPT-4 & 4.318 & 2.974 & 3.113 & 279 & <0.01 \\
        % \cmidrule[lr]{1-7}
        % {\SetCell[r=5]{l}{\textbf{Engagement**}}}
        % & Human & 4.521 & 1.064 \\
        % & GPT-4 & 4.371 & 1.117 & 3.163 & 279 & <0.01 \\
        % & T5-Large fine-tuned & 4.121 & 1.179 & 6.507 & 279 & <0.01 \\
        \bottomrule
    \end{booktabs}
    \\
    
    \caption{The paired sample t-test result of children education experts in comparison of GPT-4 and T5-Large fine-tuned on \datasetname in the QAG task. \textbf{Bolded numbers} are the best scores within each dimension excluding human experts' annotations. * means p-value <0.05, and ** means p-value <0.01, both are statistically significant.}
    \label{tab:human_eval}
    \vspace{-\baselineskip}
\end{table*}

To further compensate for the limitation of Rouge-L and \textbf{SBERT} as well as to thoroughly assess the quality and usability of LLM-generated QA pairs, particularly in terms of educational appropriateness, we conducted a human study with four education experts to compare expert-annotated QA pairs and those generated by fine-tuned T5-Large and GPT-4 with 5-shot ICL, the best-performing ones in automated evaluation.

% More specifically, according to the superior performance of fine-tuned T5-Large and GPT-4 with 5-shot ICL approach, we select these two models along with experts' annotation for human evaluation.

We randomly select ten story books from the test split of \datasetname, and sample seven sections per book.
For each section, three QA pairs are created based on the story narrative (experts' annotations, and QA pairs generated by GPT-4 and fine-tuned T5-Large), summing up $210$ QA pairs for the human evaluation. 
QA pairs are randomized for each section, and the sources are omitted to the human subjects for a fair evaluation. 

Four experts evaluate each QA pair on the following four dimensions with a 5-point Likert scale: 
\begin{enumerate}
    \item \textit{Grammar Correctness}: The QA pair uses comprehensible English Grammar;
    \item \textit{Answer Relevancy}: The answer is correct and corresponds to a question;
    \item \textit{Contextual Consistency}: The QA pair originates from the story and goes beyond the story's immediate context;
    \item \textit{Children's Educational Appropriateness}: The QA pair is appropriate for children’s reading experience during interactive story reading.
    %\item \textbf{Engagement}: The generated QA pair can capture the interest of children.
\end{enumerate}

\subsubsection{Results and Analysis}

Table~\ref{tab:human_eval} illustrates the average scores of each dimension and paired sample \textit{t-test} results.
We observe that expert-created QA pairs outperform those generated by models in all four dimensions.
The paired sample \textit{t-test} results show that experts' annotations are significantly different in three out of four dimensions compared with models' generation.
These justify \datasetname's utility in catering to teachers' real-world needs in interactive story reading.

In terms of \textit{Grammar Correctness} and \textit{Answer Relevancy}, GPT-4 achieves better performance than the fine-tuned T5-Large.
We believe it to be reasonable because LLMs such as GPT-4 are trained on vast amounts of corpora, enabling them to generate QA pairs with greater consistency in word usage. 
Therefore, compared with T5-Large, GPT-4 produces answers that connect more closely with the questions, resulting in greater coherence and accuracy between the questions and the answers.

In terms of \textit{Contextual Consistency}, the fine-tuned T5-Large significantly outperformed GPT-4, behind experts' annotations. A similar result could be found in \textit{Children's Educational Appropriateness}, wherein the T5-Large model fine-tuned on \datasetname also exhibits better performance.

These results suggest that fine-tuned with experts' annotations, the T5-Large model can generate QA pairs that 1) contain external structured knowledge connected to the story narrative, and 2) are appropriate for young children to learn during the interactive story reading activities.

\subsection{Discussion}

%% triple from KG --> QA: future work on Real-World KG-Augmented task.

% OBSERVATION 1 Finetuning with T5-Large reaches better performance --> our dataset's utility
Comparing the best-performing SoTA LLMs in the QAG pipeline with the corresponding fine-tuned T5-Large, we can observe that the T5-Large can reliably generate QA pairs aligned more with experts' annotations in terms of Rouge-L score according to system evaluation, regardless of whether generating QA pairs along real-world knowledge triples.
Drawing from the results of our human evaluation, the fine-tuned T5-Large exhibits better capabilities in generating QA pairs that suit teachers' real-world educational expectations of interactive story reading: originating from the story and embodying educational-appropriate real-world knowledge.
Worth mentioning that T5-Large only consists of $770$ million parameters, whereas Alpaca-7B, Mistral-7B, and Llama 2 in our experiments consist of $7$ billion parameters ($10$ times larger). 

This observation justifies \datasetname's utility in training a task-specific model that caters to teachers' real-world story reading needs on the one hand, and \textbf{demonstrates the usefulness of combining structured real-world knowledge and free-form narratives in domain-specific tasks such as interactive story reading.}
% FLAN-T5-XXL consists of $11$ billion parameters ($14$ times larger). 

\section{Conclusion and Future Work}

In summary, we propose \datasetname, an expert-annotated, external-knowledge-enriched QA dataset for children education, by leveraging a novel annotation framework to facilitate scalable expert annotations through structured external knowledge.
We demonstrate the effectiveness of \datasetname through an automated evaluation on various LLMs of generating QA pairs catering to teachers' needs and a human evaluation with children education experts.

One possible future work is refining the QAG pipeline structures and exploiting LLMs to generate QA pairs that align more closely with teachers' practical needs. Another future direction involves using \datasetname and language models to develop a human-AI collaborative education system (e.g., an interactive story reading system)~\cite{wang2020human,wang2019human}, aiding parents and educators to formulate personalized questions during story readings, while addressing their language, knowledge, or time constraints.
Also, fine-tuning LLMs (e.g., Llama 3) may lead to better performance on the QAG task, which offers a future direction to refine models' capabilities in real-world tasks like children's education.

% \section{Acknowledgement}

% We thank the Lab of Artificial Intelligence for Education, East China Normal University, Shanghai Institute of Artificial Intelligence for Education, East China Normal University, and Shanghai Science and Technology Innovation Action Plan project (No. 21511104500) for funding this research.  

\section{Limitations}

This work primarily focuses on constructing an expert-annotated, large-scale QA dataset consisting of story-based QA pairs associated with real-world knowledge beyond the story narrative, however, this work is not without limitations.
We could further explore LLMs' QAG capabilities with different models (e.g., GPT-4 and Llama 3), and domain-specific prompting methodologies, such as ICL with more demonstrations and RAG approaches~\cite{edge2024local} with multi-step generation pipelines.
In addition, the size and scope limitations of expert annotations in our dataset may not be sufficient for developing NLP technologies that can be generalizable to similar scenarios. 
We call for future research to explore methods for scaling the data annotation process in real-world settings and to investigate strategies for efficiently optimizing or evaluating NLP technologies in low-resource scenarios where expert resources are scarce.

\bibliography{custom}

\clearpage

\appendix

\section{Appendix}
\label{sec:appendix}

\subsection{Sample Data of \datasetname}
\label{app: sample}
In Table~\ref{tab:example_gt_qa1} and ~\ref{tab:example_gt_qa2}, we present the sample data from \datasetname, which include expert-selected concepts, real-world knowledge triples, and created QA pairs.

\begin{table}[h!]
    \centering
    \begin{booktabs}{
        width=\linewidth,
        colspec={Q[l,co=1,m]},
    }
    \toprule[1.2pt]
    \textcolor{c_blue}{\textbf{Story Section:}}\\
     {
     At the time when the Tang dynasty reigned over the Middle Kingdom, there were master swordsmen of various kinds. \\
     Those who came first were the saints of the sword. They were able to take different shapes at will, and their swords were like strokes of lightning.\\
     ...\\
     They wore a hidden \textbf{{\textcolor{c_green}{\underline {dagger}}}} at their side and carried a leather \textbf{{\textcolor{c_green}{\underline {bag}}}} at their belt. \\
     By magic means they were able to turn human heads into flowing water. \\
     ... }\\
    \midrule
    % \midrule
    \textbf{Expert annotated QA pairs} \\
    \midrule
    % \midrule
    \textbf{Triple: } {(\textbf{{\textcolor{c_green}{\underline {dagger}}}}, \textit{\textbf{is a}}, \textcolor{c_orange}{short sword})}\\
    \textbf{Question: }\textit{\textbf{What is}} a \textcolor{c_orange}{short sword} called? \\
    \textbf{Answer: }A \textbf{{\textcolor{c_green}{\underline {dagger}}}}.\\
    \midrule
    \textbf{Triple: } {(\textbf{{\textcolor{c_green}{\underline {bag}}}}, \textit{\textbf{is used for}}, \textcolor{c_orange}{carrying things})}\\
    \textbf{Question: }\textit{\textbf{What is}} a \textbf{{\textcolor{c_green}{\underline {bag}}}} \textit{\textbf{used for}}? \\
    \textbf{Answer: }A bag is used for \textcolor{c_orange}{carrying things}.\\
    \bottomrule[1.2pt]
    \end{booktabs}
    \caption{Example 1 of expert annotated data point in \datasetname.}
    \label{tab:example_gt_qa1}
\end{table}

\begin{table}[h!]
    \centering
    \begin{booktabs}{
        width=\linewidth,
        colspec={Q[l,co=1,m]},
    }
    \toprule[1.2pt]
    \textcolor{c_blue}{\textbf{Story Section:}}\\
     {...\\
     On hearing this the king walked to the window and stood for a few minutes with his back to the room, where the company of young men remained silent. Then he came back, his face \textbf{{\textcolor{c_green}{\underline {white}}}} and stern.\\
     'I tell you,' he said, 'and it is the solemn truth, that I would rather you had told me that the prince was dead, though he is my only son, than know that he would suffer such an \textbf{{\textcolor{c_green}{\underline {injury}}}} without attempting to avenge it\\
     ...}\\
    \midrule
    % \midrule
    \textbf{Expert annotated QA pairs} \\
    \midrule
    % \midrule
    \textbf{Triple: } {(\textbf{{\textcolor{c_green}{\underline {white}}}}, \textit{\textbf{is a}}, \textcolor{c_orange}{color})}\\
    \textbf{Question: }\textit{\textbf{What}} \textcolor{c_orange}{color} is snow? \\
    \textbf{Answer: }A \textbf{{\textcolor{c_green}{\underline {White}}}}.\\
    \midrule
    \textbf{Triple: } {(\textbf{{\textcolor{c_green}{\underline {injury}}}}, \textit{\textbf{is at location of}}, \textcolor{c_orange}{hospital})}\\
    \textbf{Question: }\textit{\textbf{Where}} do you go if you get very \textbf{{\textcolor{c_green}{\underline {hurt}}}}? \\
    \textbf{Answer: }You go to \textcolor{c_orange}{hospital} if you get very hurt.\\
    \bottomrule[1.2pt]
    \end{booktabs}
    \caption{Example 2 of expert annotated data point in \datasetname.}
    \label{tab:example_gt_qa2}
\end{table}

\subsection{Properties of Educational QA datasets}
\label{EduQADataset}

The key properties of educational QA datasets, including their number and type of sourcebooks, QA pairs, whether they contain external knowledge, annotators, annotation process, and data composition, are presented in Table~\ref{tab:related_datasets}.

\begin{table*}[h!]
    \centering
    \resizebox{.99\linewidth}{!}{
    \begin{tabular}{lccccccc}
        \toprule[1pt]
        \textbf{Dataset} & \textbf{\# Books} & \textbf{\# QA Pairs} & \textbf{\makecell[c]{External\\Knowledge}} & \textbf{Annotator} & \textbf{\makecell[c]{Document\\Source}} & \textbf{\makecell[c]{Annotation\\Process}} & \textbf{\makecell[c]{Data\\Composition}} \\
        \midrule
        \midrule
        StoryQA & 148 & 38,703 & Yes & Crowd-Sourced & Story books & \makecell[c]{Direct\\Annotation} & \makecell[l]{1. Story Section\\2. QA Pairs} \\
        % \cdashline{1-8}
        \midrule
        FairytaleQA & 278 & 10,580 & No & Expert & Story books & \makecell[c]{Direct\\Annotation} & \makecell[l]{1. Story Section\\2. QA Pairs} \\
        % \cdashline{1-8}
        \midrule
        EduQG & 13 & 5,018 & No & Expert & Text books & \makecell[c]{Direct\\Annotation} & \makecell[l]{1. Source Documents\\2. Questions and Answer Options} \\
        \midrule
        \datasetname & 278 & 5,868 & Yes & Expert & Story books & \makecell[c]{3-step Guided\\Annotation} & \makecell[l]{1. Story Section\\2. Real-world Knowledge Triples\\3. QA Pairs} \\
        \bottomrule[1pt]
    \end{tabular}
    }
    \caption{Properties of existing datasets focusing on children education compared with our \datasetname.}
    \label{tab:related_datasets}
\end{table*}

\subsection{ConceptNet Relations}
\label{app:filter_relations}

We follow \citet{xuFusingContextKnowledge2020}'s work to filter out weak relations in ConceptNet, and our ranking algorithm uses the following 13 relations in our annotation framework as well as GPT prompts:
\textit{causes}, \textit{desires}, \textit{has context of}, \textit{has property}, \textit{has subevent}, \textit{is a}, \textit{is at location of}, \textit{is capable of}, \textit{is created by}, \textit{is made of}, \textit{is part of}, \textit{is the antonym of}, \textit{is used for}.

\subsection{Distribution of Question Type}
\label{app:distribution}

The distribution of question type in \datasetname is shown in Table~\ref{tab:distribution}.

\begin{table}[h!]
    \centering
    \begin{booktabs}{
        colspec={ccccr},
        row{1}={font=\bfseries\small},
        cells={m},
    }
    \toprule
     Interrogative & {Train\\Split} & {Val\\Split} & {Test\\Split} & \SetCell[c=1]{c}{Total\\Percentage (\%)}\\
     \midrule
     \midrule
    what & 3779 & 628 & 641 & 86.01 \\
    why & 227 & 93 & 105 & 7.24 \\
    who & 76 & 10 & 14 & 1.70 \\
    where & 41 & 3 & 7 & 0.87\\
    when & 20 & 12 & 8 & 0.68\\
    how & 112 & 13 & 15 & 2.39\\
    other & 42 & 10 & 9 & 1.04\\
    \bottomrule
    \end{booktabs}
    \caption{Distribution of question types in \datasetname.}
    \label{tab:distribution}
\end{table}

\subsection{Analysis of the Type of Generated External Knowledge}
\label{app: distribution_analysis}

We calculated the real-world knowledge type distribution of experts’ annotation and triples generated by T5-Large and GPT-4 with 5-shot ICL (the best-performing ones in automated evaluation). The results are shown in Table~\ref{tab:distribution_compare}.

According to the result, the fine-tuned T5 model can generate real-world knowledge triples more aligned with experts’ annotation, surpassing GPT-4, further proving that domain-specific fine-tuning provides the model with targeted knowledge and expertise that LLMs may lack. 

\begin{table}[h]
\centering
\resizebox{0.99\linewidth}{!}{
\begin{tabular}{lccc}
\toprule[1pt]
\textbf{Relations} & \textbf{\makecell[c]{Experts'\\Annotation}} & \textbf{\makecell[c]{T5-Large\\Fine-Tuned}} & \textbf{GPT-4} \\ 
\midrule
\midrule
\textit{is a} & 35.45\% & 47.67\% & 20.64\% \\ 
% \midrule
\textit{has subevent} & 16.21\% & 16.44\% & 4.66\% \\ % \midrule
\textit{is the antonym of} & 15.20\% & 7.95\% & 1.64\% \\ % \midrule
\textit{is used for} & 8.78\% & 9.32\% & 35.25\% \\ % \midrule
\textit{is at location of} & 7.53\% & 7.40\% & 2.19\% \\ % \midrule
\textit{is capable of} & 5.18\% & 4.11\% & 9.77\% \\ % \midrule
\textit{other} & 11.65\% & 6.85\% & 25.84\% \\ 
\bottomrule[1pt]
\end{tabular}
}
\caption{Comparison of real-world knowledge relation types across experts' annotations, fine-tuned T5-Large, and GPT-4.}
\label{tab:distribution_compare}
\vspace{-\baselineskip}
\end{table}

\subsection{Hyper-Parameters and Experiment Settings}
\label{app:hyper}
We conducted our experiments on Google Colab with A100.
Following common practice when fine-tuning the T5-Large model, we use the learning rate of 1e-4 and train our model on 3 epochs.

\subsection{Complete QAG Pipeline Results}
\label{app:FullResult}
We demonstrate the complete performance of LLMs in our QAG pipeline using both zero-shot and few-shot ICL approaches in Table~\ref{tab:end2end_full_result}.

\begin{table}[h]
    \centering
    \resizebox{0.99\linewidth}{!}{
    \begin{booktabs}{
    colspec={cl| cccc},
    row{1}={font=\bfseries\small},
    width=\linewidth,
    cells={m},
}
\toprule[1pt]
\SetCell[r=2]{c}{Models} & \SetCell[r=2]{c}{Prompting\\Strategy} & \SetCell[c=2]{c}{End2End Pipeline\\w/o Triples} && \SetCell[c=2]{c}{End2End Pipeline\\w/ Triples}\\
&& Rouge-L & {SBERT} & Rouge-L & {SBERT} \\
\midrule
\midrule
{T5-Large\\Fine-Tuned (0.77B)} & - & \textbf{0.332} & 0.289 & \textbf{0.279} & 0.263 \\
\midrule
\SetCell[r=2]{c}{Alpaca\\(7B)} & zero-shot & 0.124 & 0.186 & 0.266 & 0.207 \\
& 1-shot & 0.251 & 0.182 & 0.239 & 0.186 \\
\midrule
\SetCell[r=3]{c}{Mistral\\(7B)} & zero-shot & 0.229 & 0.237 & 0.209 & 0.229 \\
& 1-shot & 0.227 & 0.237 & 0.231 & 0.241\\
& 5-shot & 0.267 & 0.241 & 0.257 & 0.251\\
\midrule
\SetCell[r=3]{c}{Llama 2\\(7B)} & zero-shot & 0.213 & 0.234 & 0.177 & 0.225 \\
& 1-shot & 0.192 & 0.217 & 0.206 & 0.237 \\
& 5-shot & 0.241 & 0.240 & 0.269 & 0.253 \\
\midrule
{Flan-T5-XXL} & 1-shot & 0.264 & 0.246 & 0.194 & 0.209 \\
\midrule
\SetCell[r=4]{c}{GPT-3.5} & zero-shot &  0.194 & 0.233 & 0.220 & 0.252\\
& 1-shot & 0.239 & 0.262 & 0.252 & 0.271 \\
& 5-shot & 0.262 & 0.279 & 0.264 & 0.266 \\
& CoT & - & - & 0.259 & 0.280 \\
\midrule
\SetCell[r=4]{c}{GPT-4} & zero-shot &  0.277 & 0.252 & 0.243 & 0.261 \\
& 1-shot & 0.272 & 0.279 & 0.251 & \textbf{0.292} \\
& 5-shot & 0.287 & \textbf{0.311} & 0.248 & 0.283 \\
& CoT & - & - & 0.262 & \textbf{0.292} \\
\bottomrule[1pt]
\end{booktabs}}
\caption{
 Rouge-L and SentenceBERT scores of LLMs in the QAG task. \textbf{Bolded numbers} are global best performance within each setting on each metric.
}
\label{tab:end2end_full_result}
\vspace{-\baselineskip}
\end{table}

\subsection{LLM Prompts}
\label{app: gpt_prompts}
To utilize LLMs' strong reasoning and generation capability as well as control GPT-generated questions as much as possible to meet the needs of teachers, we carefully design our prompts.

For the QAG pipeline, there are two variations based on the system:
(1) Directly generate a QA pair based on a provided story section.
(2) From a story section, generate a real-world knowledge triple and a QA pair simultaneously.

Table~\ref{tab:gpt_prompt_1},~\ref{tab:gpt_prompt_2} list our prompts for GPT in the two abovementioned approaches.

\begin{table*}[!h]
\centering
\begin{booktabs}{
    width=\linewidth,
    colspec={Q[l,co=1,m]},
    cell{1}{1}={c},
} 
\toprule
{
\textbf{Prompt for LLMs in the QAG Pipeline}\\
\textbf{(Generate QA Pairs Only)}
} \\
\midrule
\midrule
I need you to help generate a question and answer pair for young children aged three to six. I will provide you with a short section of a story delimited by triple quotes.
Please follow these steps:\\
1. For each sentence, identify one key word that meets the following criteria: it is relatively complex, it is considered tier 1 or tier 2 vocabulary, and it is a concrete noun, verb, or adjective.\\
2. After this, you need to completely forget about the story that I gave you, remembering only the words you identified.\\
3. Based on each selected word, generate a question and answer pair that either the question or the answer contains that word. For example, if your identified word is 'apple', your question could be: where do apples grow? what do apples taste like? what color are apples? These questions should go beyond the context of the stories. \\
Each question should have one single correct answer that would be the same regardless of the children's experiences. 
The questions should be focused on real-world, fact-based knowledge and beneficial to educate children during story reading. \\
The real-world, fact-based knowledge should be based on the selected word and is in the form of a triple such as A relation B, where A and B are two concepts and the selected word can be either A or B. You should use one of the following relations for the real-world knowledge: \\
  ~~~~1) causes, 2) desires, 3) has context of, 4) has property, 5) has subevent, 6) is a, 7) is at location of, \\
  ~~~~8) is capable of, 9) is created by, 10
  ) is made of, 11) is part of, 12) is the antonym of, 13) is used for\\
4. After this, select one question-answer pair that you think best meets my criteria. Please note that the question should be answerable without reading the story. The answer should only be a concrete noun, verb, or adjective.\\
Return the selected question-answer pair in the following format: \\
\\
%\hfill

question: ...\\
answer: ...\\
\\

\textlangle story\textrangle: \\
\textbf{\textit{ \{story1 for few-shot\} }}\\
\\

\textlangle response\textrangle: \\
\textbf{\textit{ \{response1 for few-shot\}}}\\

... ...\\

\\
\textlangle story\textrangle: \\
\textbf{\textit{ \{story for the current data\}  } }\\
\\
\textlangle response\textrangle: \\

\bottomrule
\end{booktabs}
\caption{ Prompt for LLMs in the QAG task with generating QA pairs directly from the story. }
\label{tab:gpt_prompt_1}
\end{table*}

\begin{table*}[!h]
\centering
\resizebox{0.99\linewidth}{!}{
\begin{booktabs}{
    width=\linewidth,
    colspec={Q[l,co=1,m]},
    cell{1}{1}={c},
} 
\toprule
{
\textbf{Prompt for LLMs in the QAG Pipeline}\\
\textbf{(Generate Triples and QA Pairs)}
} \\
\midrule
\midrule
I need you to help generate a question and answer pair for young children aged three to six. I will provide you with a short section of a story delimited by triple quotes.
Please follow these steps:\\
1. For each sentence, identify one key word that meets the following criteria: it is relatively complex, it is considered tier 1 or tier 2 vocabulary, and it is a concrete noun, verb, or adjective.\\
2. After this, you need to completely forget about the story that I gave you, remembering only the words you identified.\\
3. Based on each selected word, generate one real-world relation based on the selected word. This real-world relation should go beyond the context of the stories. For example, if your identified word is ’apple’, your real-world relation could be: apple grows on trees; apples are red. The real-world, fact-based knowledge should be based on the selected word and is in the form of a triple such as 'A relation B', where A and B are two concepts and the selected word can be either A or B. 
You should use one of the following relations for the real-world knowledge:  \\
 ~~~~1) causes, 2) desires, 3) has context of, 4) has property, 5) has subevent, 6) is a, 7) is at location of, \\
  ~~~~8) is capable of, 9) is created by, 10
  ) is made of, 11) is part of, 12) is the antonym of, 13) is used for\\
4. After this, generate a question and answer pair based on the real-world, fact-based knowledge you generated. Either the question or the answer should contain that identified word. Each question should have one single correct answer that would be the same regardless of the children's experiences. The questions should be focused on real-world, fact-based knowledge and beneficial to educate children during story reading.\\
5. After this, select one question-answer pair that you think best meets my criteria. Please note that the question should be answerable without reading the story. The answer should only be a concrete noun, verb, or adjective.\\
Return the generated real-world knowledge triple and selected question-answer pair in the following format: \\
\\
real-world knowledge triple: (A, relation, B)\\
question: ...\\
answer: ...\\

\hfill

\textlangle story\textrangle: \\
\textbf{\textit{ \{story1 for few-shot\} }}\\
\hfill

\textlangle response\textrangle: \\
\textbf{\textit{ \{response1 for few-shot\}}}\\

... ...\\

\hfill

\textlangle story\textrangle: \\
\textbf{\textit{ \{story for the current data\} } } \\
\hfill

\textlangle response\textrangle: \\

\bottomrule
\end{booktabs}
}
\caption{ Prompt for LLMs in the QAG task with generating real-world knowledge triple and QA pairs directly from the story. }
\label{tab:gpt_prompt_2}
\end{table*}

\subsection{User Interface for Annotation System}

\begin{figure*}[ht]
    \centering
    \includegraphics[width=\linewidth]{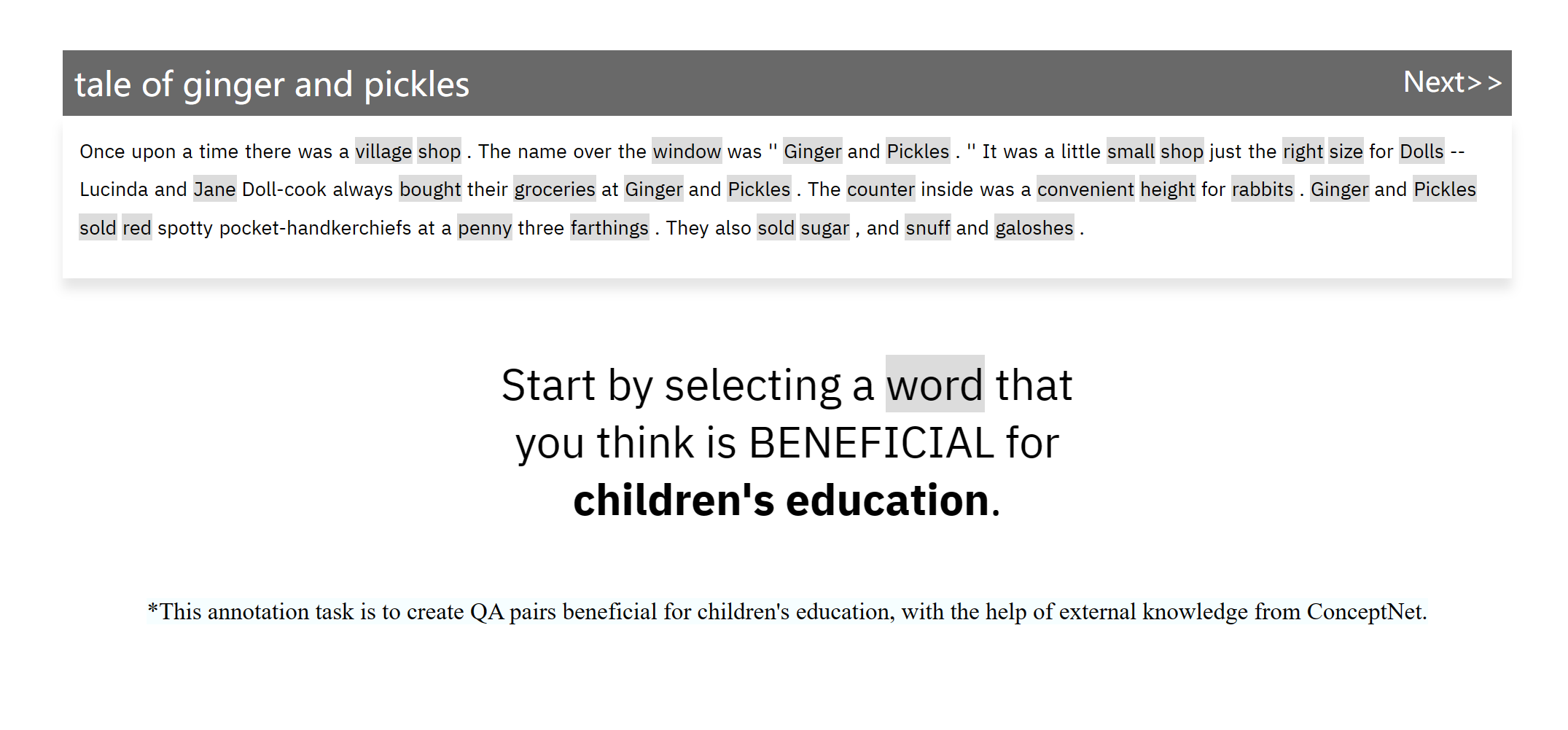}
    \caption{Annotation process 1: browse a displayed section, with candidate words highlighted in grey.}
    \label{fig:section}
\end{figure*}

\begin{figure*}[ht]
    \centering
    \includegraphics[width=\linewidth]{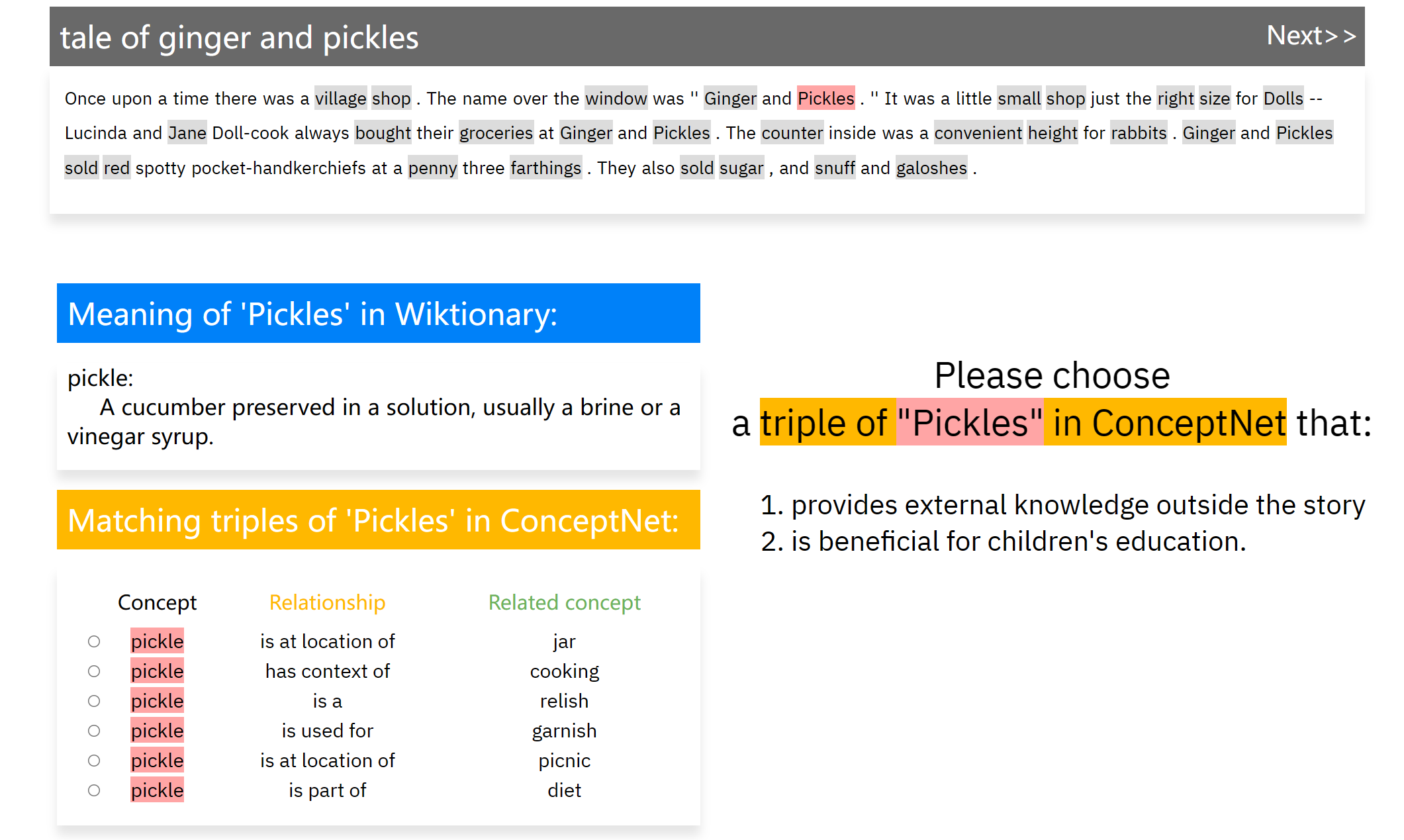}
    \caption{Annotation process 2: after selecting a \textcolor{red}{word} (highlighted in red), related explanation in Wiktionary and candidate real-world knowledge triples in ConceptNet will display.}
    \label{fig:triples}
\end{figure*}

We implement an annotation system to facilitate QA pair annotation with associated external knowledge.
Figure~\ref{fig:section},~\ref{fig:triples} and~\ref{fig:ui_QA_creation} show the annotation interface for human experts.

\begin{figure*}[ht]
    \centering
    \includegraphics[width=\linewidth]{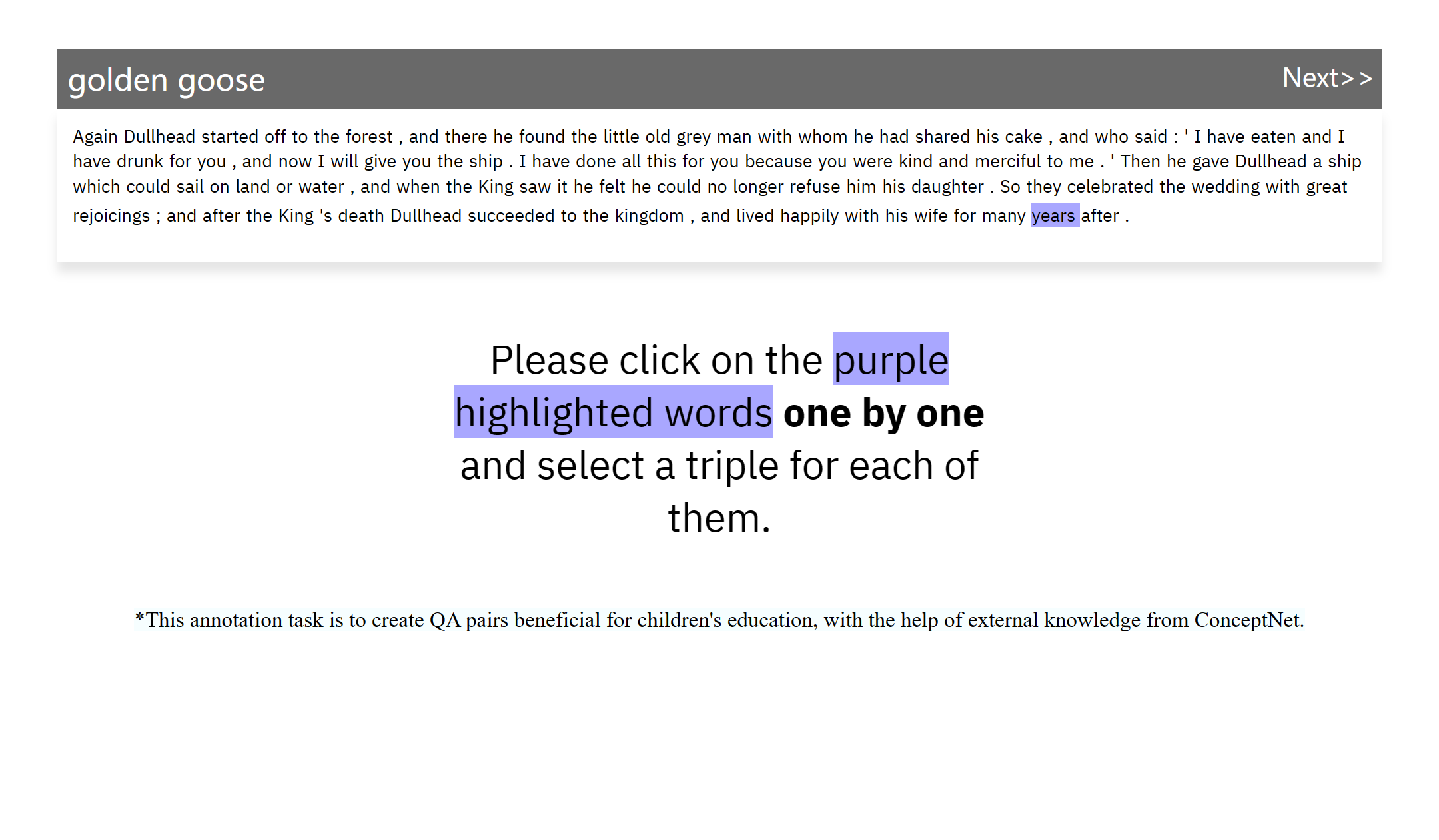}
    \caption{Cross-validation process 1: browse a displayed section, with candidate words highlighted in grey.}
    \label{fig:val-section}
\end{figure*}

\begin{figure*}[ht]
    \centering
    \includegraphics[width=\linewidth]{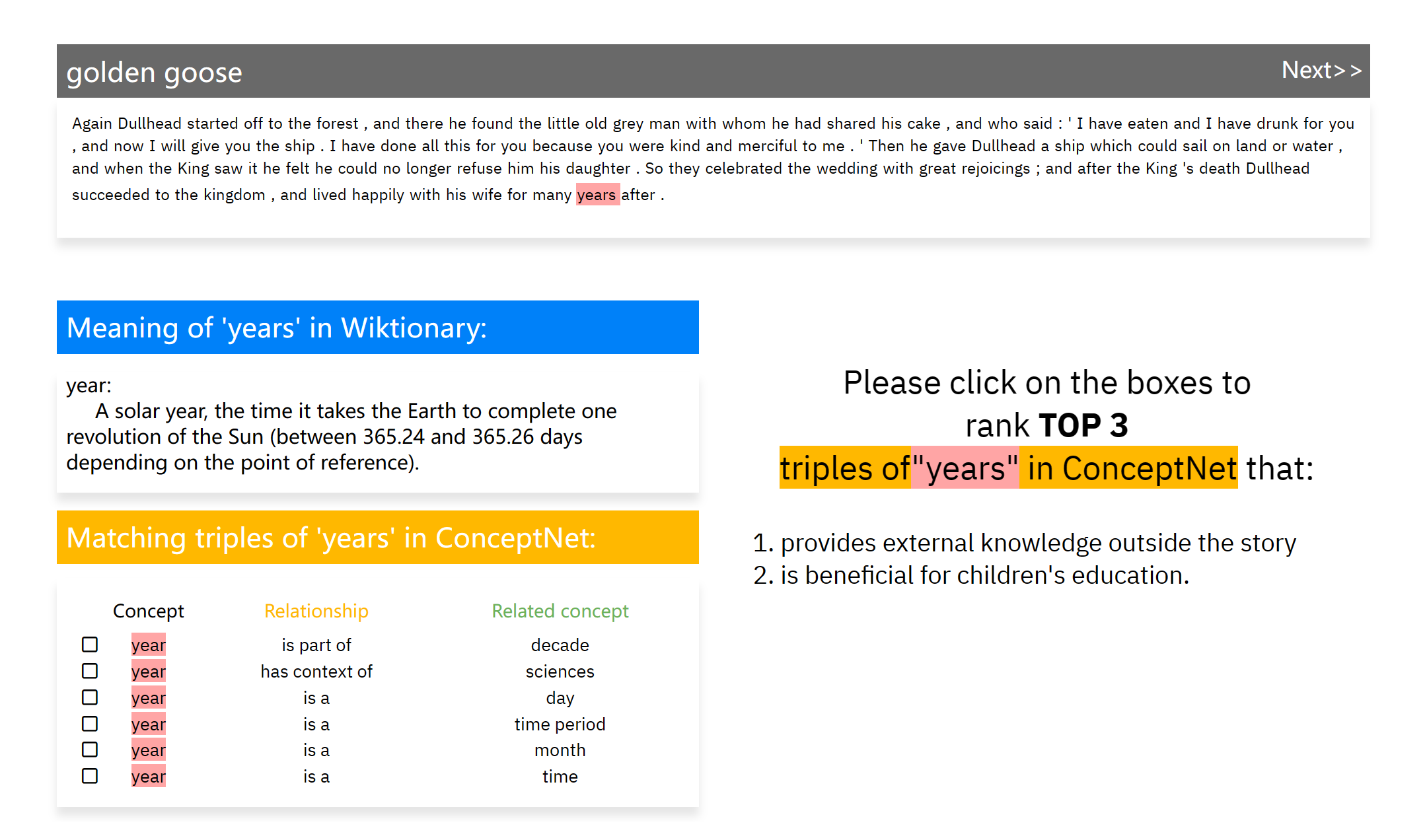}
    \caption{Cross-validation process 2: select a word annotated by others and rank the candidate triples.}
    \label{fig:val-rank}
\end{figure*}

\begin{figure*}[ht]
    \centering
    \includegraphics[width=\linewidth]{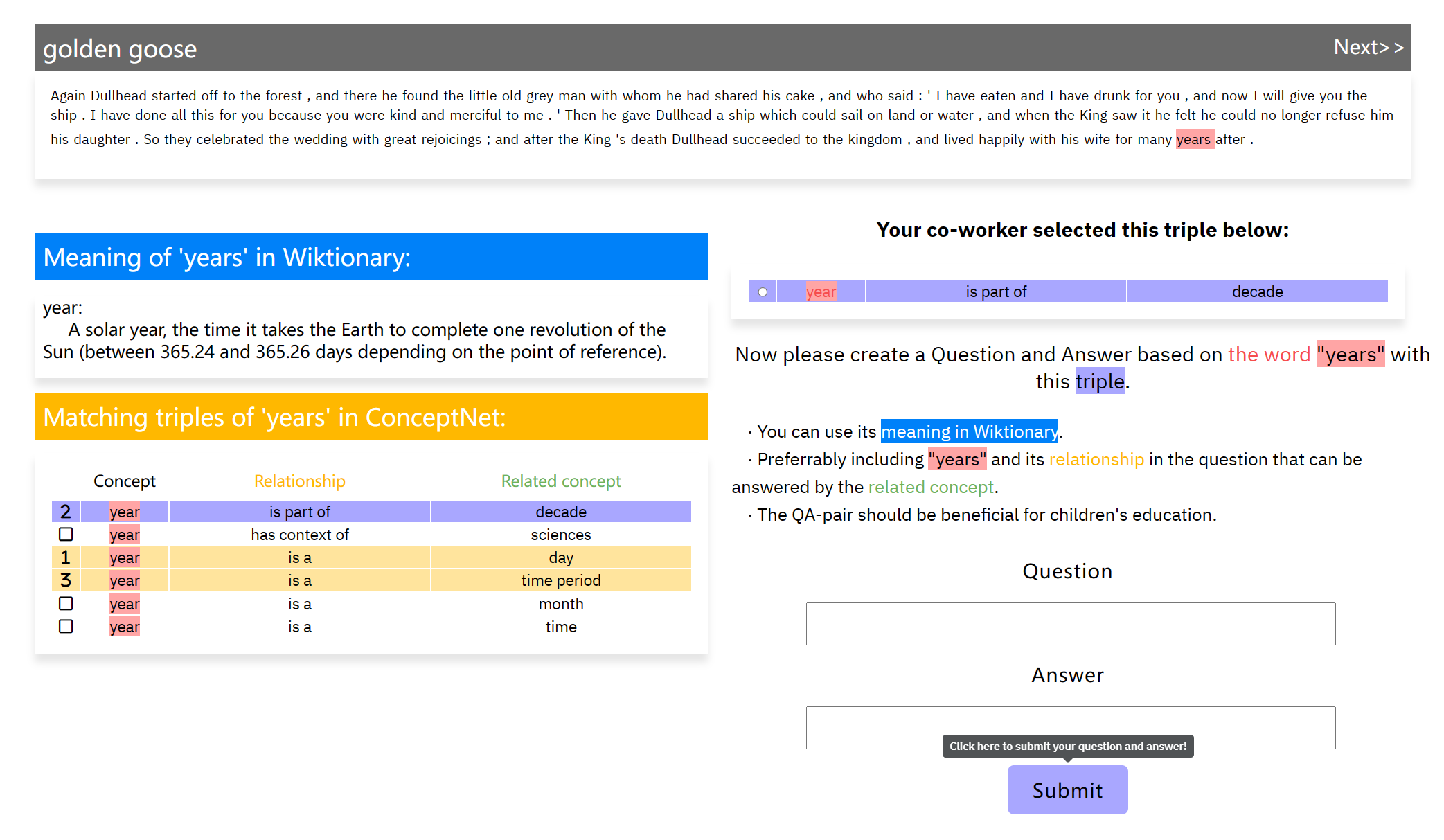}
    \caption{Cross-validation process 3: after ranking top3 triples, the triple selected originally by the other annotator is displayed, the validator should create a QA pair based on the original triple.}
    \label{fig:val-QA}
\end{figure*}

\begin{figure*}[ht]
    \centering
    \includegraphics[width=\linewidth]{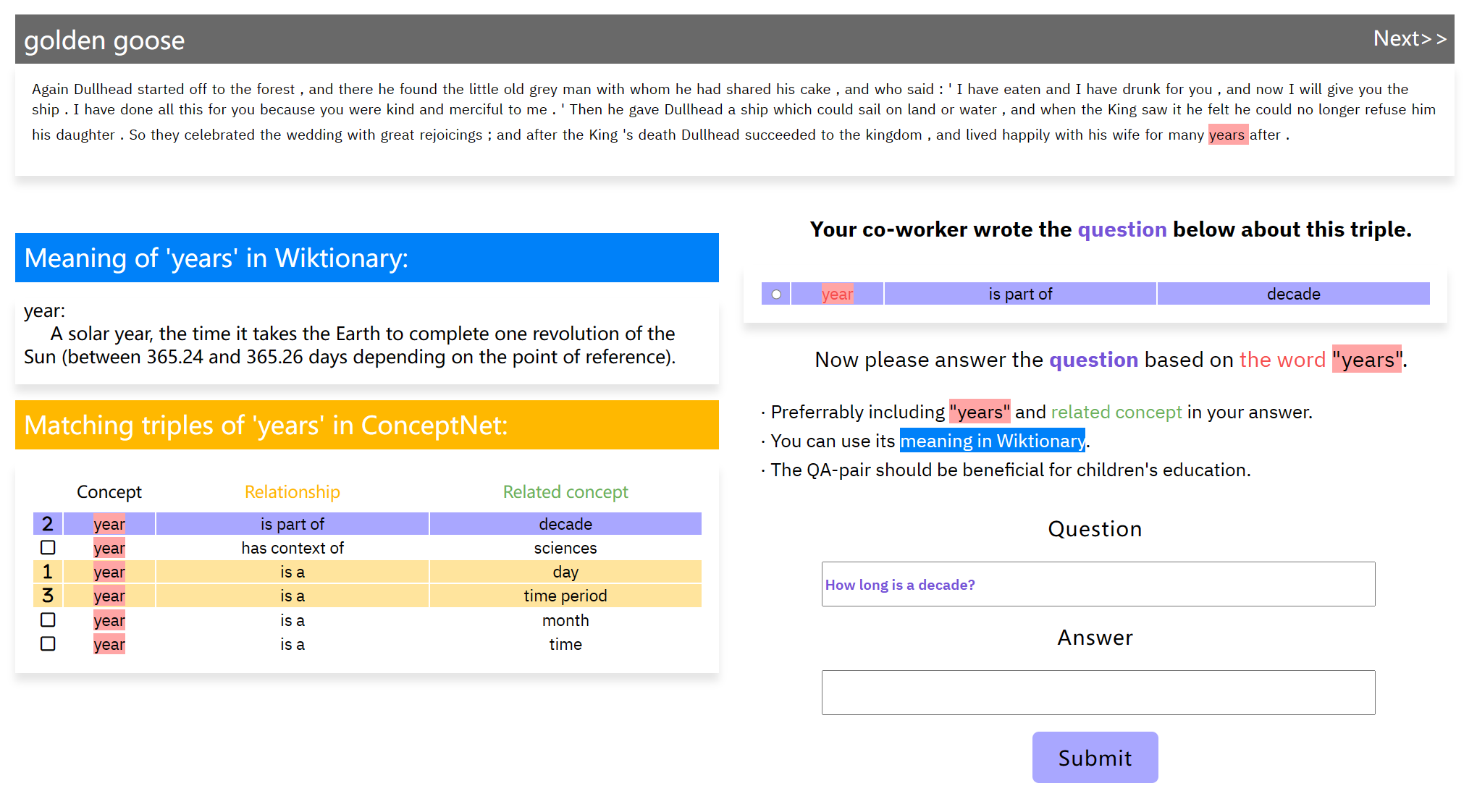}
    \caption{Cross-validation process 4: validator is asked to answer the question created by the other annotator using the triple originally selected by the other annotator.}
    \label{fig:val-answer}
\end{figure*}

We also conduct cross-validation to assess the agreement among annotators.
Figure~\ref{fig:val-section},~\ref{fig:val-rank},~\ref{fig:val-QA} and~\ref{fig:val-answer} demonstrate user interfaces for each step to support the cross-validation process.

\end{document}